\documentclass{verts}
\usepackage[utf8]{inputenc}
\usepackage[T1]{fontenc}

\usepackage{bm}
\usepackage{scalerel}
\usepackage[inference]{semantic}

\usepackage{svg}
\usepackage{tikz}
\usetikzlibrary{cd}

\usepackage[draft,author=MS]{fixme}
\fxsetup{layout=inline,marginclue}
\fxsetup{theme=color}

\providecommand\given{}
\providecommand\and{}

\newcommand\SetSymbol[1][]{%
	\nonscript\:#1\vert%
	\allowbreak\nonscript\:%
	\mathopen{}}

\DeclarePairedDelimiterX\set[1]\{\}{%
	\renewcommand\given{\SetSymbol[\delimsize]}
	\def\and{\:\land\:}
	\, #1 \,
}

\DeclarePairedDelimiterX
{\norm}[1]
{\lVert}{\rVert}
{\ifblank{#1}{\,\cdot\,}{#1}}

\DeclarePairedDelimiterX
{\abs}[1]
{\lvert}{\rvert}
{\ifblank{#1}{\,\cdot\,}{#1}}

\DeclarePairedDelimiterX
{\sem}[1]
{\llbracket}{\rrbracket}
{\ifblank{#1}{\,\cdot\,}{#1}}

\DeclarePairedDelimiterX
{\innerprod}[2]
{\langle}{\rangle}
{\ifblank{#1}{\,\cdot\,}{#1},\ifblank{#2}{\,\cdot\,}{#2}}

\DeclarePairedDelimiterX
{\repr}[1]
{[}{]}
{\ifblank{#1}{\,\cdot\,}{#1}}

\DeclareMathOperator{\Relu}{ReLU}
\DeclareMathOperator{\idfunc}{id}

\DeclareMathOperator{\true}{tt}

\DeclareMathOperator{\domain}{dom}
\DeclareMathOperator{\codomain}{codom}

\DeclareMathOperator{\bigO}{\mathcal{O}}

\newcommand{\realset}{\mathbb{R}}
\newcommand{\natset}{\mathbb{N}}
\newcommand{\binaryset}{\mathbb{B}}

\newcommand{\plnnset}{\mathcal{N}}
\newcommand{\affset}{\Phi}
\newcommand{\tadsset}{\Theta}

\newcommand{\plnn}{N}
\newcommand{\aff}{\alpha}
\newcommand{\paff}{\psi}
\newcommand{\relu}[2]{\ifblank{#1}{\phi_{#2}}{\phi^{#1}_{#2}}}

\newcommand{\defined}{\mathrel{\vcentcolon=}}

\newcommand{\indicatorfunc}{\mathbb{I}}

\newcommand{\inn}{n}
\newcommand{\out}{m}

\newcommand{\mat}[1]{\bm{#1}}
\newcommand\idmat[1]{\mat{I}^{#1}}
\newcommand\idzmat[2]{\mat{E}^{#1}_{#2}}

\newcommand{\dotprod}[2]{{{#1}^\top \!} \cdot {#2}}
\newcommand{\unitvec}[1]{\vec{\mathrm{e}_{#1}}}

\newcommand{\config}[1]{\langle {#1} \rangle}
\newcommand{\sosto}[1]{\xrightharpoonup{#1}}
\newcommand{\symderiv}[1]{\sosto{#1}_{\scriptscriptstyle\mathrm{SOS}}}
\newcommand{\pc}{\pi}

\newcommand\bnfeq{\:\mathrel{\vcentcolon\vcentcolon=}\:}
\newcommand\bnfmid{\,\mathbin{\vert}\,}

\newcommand{\nettransform}{\tau}
\newcommand\netcomp{\mathbin{;}}
\newcommand\nettotads{\tau}
\newcommand{\netsem}[1]{\sem{#1}_{\scriptscriptstyle\mathrm{DS}}}

\newcommand{\algebra}[1]{#1}
\newcommand{\algcarrier}[1]{\mathcal{#1}}
\newcommand{\algoperations}[1]{#1}
\newcommand{\algebraintro}[3]{\algebra{#1} = (\algcarrier{#2},\algoperations{#3})}

\newcommand{\adsset}[1]{\mathcal{S}_{\scriptsize\algebra{#1}}}

\newcommand{\adssem}[2][A]{\sem{#2}_{\scriptscriptstyle\mathcal{S}\!_{\algebra{#1}}}}

\newcommand{\tadssem}{\semtads}
\newcommand{\tadsjoin}{\Join}

\newcommand{\semds}[1]{\sem{#1}_{\scriptscriptstyle\mathrm{DS}}}

\newcommand{\semos}[1]{\sem{#1}_{\scriptscriptstyle\mathrm{OS}}}
\newcommand{\ssemos}[1]{\sem{#1}_{\scriptscriptstyle\mathrm{SOS}}}

\newcommand{\semtads}[1]{\sem{#1}_{\scriptscriptstyle\mathrm{\tadsset}}}

\newcommand{\genericoperator}{\mathbin{\scalerel*{\square}{\diamond}}}
\newcommand{\liftedgenericoperator}{\mathbin{\scalerel*{\blacksquare}{\diamond}}}
\newcommand\liftedplus{\oplus}
\newcommand\liftedminus{\ominus}
\newcommand\liftedsmul{\odot}
\newcommand\liftedeq{\textcircled{=}}

\begin{document}

\title{Towards Rigorous Understanding of Neural Networks via Semantics-preserving Transformations}

\titlerunning{Towards Rigorous Understanding}
\authorrunning{M. Schlüter et al.}

\author{
	Maximilian Schlüter%
	\setorcid{0000-0002-5100-7259} \inst{1} \eqcontr \and
	Gerrit Nolte%
	\setorcid{0000-0002-5080-1039} \inst{1} \eqcontr \and
	Alnis Murtovi%
	\inst{1} \and
	and Bernhard Steffen%
	\setorcid{0000-0001-9619-1558} \inst{1}
}

\institute{TU Dortmund University, Dortmund, Germany}

\corresponding{maximilian.schlueter@tu-dortmund}

\abstract{
	In this paper, we present an algebraic approach to the precise and global verification and explanation of \emph{Rectifier
	Neural Networks}, a subclass of \emph{Piece-wise Linear Neural Networks} (PLNNs), i.e., networks that semantically represent piece-wise affine functions.
	Key to our approach is the symbolic execution of these networks that allows the construction 
	of semantically equivalent \emph{Typed Affine Decision Structures} (TADS). 
	Due to their deterministic and sequential nature, TADS can, similarly to
	decision trees, be considered as white-box models and therefore as precise solutions to the 
	model and outcome explanation problem.
	TADS are linear algebras, which allows one to elegantly 
	compare Rectifier Networks for equivalence or similarity, both with precise diagnostic 
	information in case of failure, and to characterize their classification potential by precisely 
	characterizing the set of inputs that are specifically classified, or the set of inputs where two network-based classifiers differ.
    All phenomena are illustrated along a detailed discussion of a minimal, illustrative example: the continuous XOR function.}


\maketitle
\section{Introduction}
Neural networks are perhaps today's most important machine learning models, with exciting results, e.g., in image recognition~\cite{simonyan2014very}, speech recognition~\cite{chiu2018state,brown2020language} and even in highly complex games~\cite{vinyals2019grandmaster,berner2019dota,silver2017mastering}.
As the name suggests, neural networks are learned from data using efficient, but approximate training algorithms~\cite{ruder2016overview,kingma2014adam}.
At their core, neural~networks are (dataflow-oriented) computation~graphs~\cite{Goodfellow-et-al-2016}. They consist of many computation units, called \textit{neurons}, that are arranged in layers such that computations in each layer can be performed in parallel, with successive layers only depending on the preceding layer. Modern neural networks, in practice, possess up to multiple billions of parameters \cite{brown2020language} and leverage parallel hardware such as GPUs to perform computations of this scale \cite{oh2004gpu}. This highly quantitative approach is responsible for exciting success stories, but also for their main weakness: Neural network behavior is often chaotic and hard to comprehend for a human. Perhaps most infamously, a neural network's prediction can change drastically under imperceptible changes to its input, so-called \emph{adversarial examples} \cite{madry2017towards,goodfellow2014explaining,szegedy2013intriguing}.

The explainability of neural networks, which are computationally considered as black-boxes due to their highly parallel and non-linear nature, is therefore one of the current core challenges in AI research \cite{doran2017does}. The fact that neural networks are increasingly used in safety-critical systems such as self-driving cars \cite{badue2021self}
turns trustworthiness of machine learning into a must \cite{doran2017does}. However, state-of-the-art explanation technology is more about reassuring intuition, e.g., to support  cooperative work of humans with AI systems, such as in the field of medical diagnostics \cite{tjoa2020survey}, than about precise explanation or guarantees \cite{linardatos2021explainable}.
Moreover, current approaches to Neural Network verification are still in their infancy in that they are not yet sufficiently tailored to the nature of Neural Networks to achieve the required scalability or to 
provide diagnostic information beyond individual witness traces 
in cases where the verification attempts fail (cf.,~\cite{bak2021second,katz2017reluplex,wang2021beta} and \cref{sec:related_work} for a more detailed discussion).

\bigskip\noindent
In this paper, we present an algebraic approach to the verification and explanation  of Rectifier Neural Networks (PLNN), a very popular subclass of neural networks that semantically 
represent piece-wise affine functions (PAF) \cite{montufar2014number}. Key to our approach are \emph{Typed Affine Decision Structures} (TADS) that concisely represent 
PAF in a white-box fashion that is as accessible to human understanding as decision trees. TADS can nicely be derived from PLNNs via symbolic execution \cite{clarke1976system,king1976symbolic}, or, alternatively, compositionally along the PLNN's layering structure, and their algebraic structure allows
for elegant solutions to verification and explanation tasks:
\begin{itemize}
	\item TADS can be used for PLNNs similarly as \emph{Algebraic Decision Diagrams} (ADDs) have been used for Random Forests in~\cite{gossen2021algebraic} to
	elegantly provide model and outcome explanations as well as class characterizations.
	\item Using the algebraic operations of TADS one can not only decide the equivalence problem, i.e., whether two PLNNs are semantically equivalent, but also whether they are $\epsilon$-similar, i.e., never differ more than
	$\epsilon$. In both cases, diagnostic information in terms of a corresponding `difference' TADS is provided that
	precisely specifies where one of these properties is violated.
	\item TADS comprise non-continuous piece-wise linear operations which cannot be represented by PLNNs. This is necessary to not only
	deal with \emph{regression tasks}, where one aims at approximating continuous functions, but also with \emph{classification tasks} with discrete output domains.\footnote{As PLNNs always represent continuous functions, an additional outcome interpretation mechanism is needed to bridge the gap from continuous networks to discrete classification tasks.}
    In the latter case, TADS-based
	class characterization allows one to precisely characterize the set of inputs that are classified as members of a given class, or the set of inputs
	where two (PLNN-based) classifiers differ.
	\item Finally, TADS can also profitably be used for the verification of  preconditions and postconditions, the illustration of which is beyond the scope of this paper, but will be discussed in~\cite{STTT4} in the setting of digit recognition.
\end{itemize}
The paper illustrates the essential features of TADS using a minimal, illustrative example: the continuous XOR function.
The simplicity of XOR is ideally suited to provide an intuitive entry into the presented theory. A more comprehensive example
is presented in~\cite{STTT4}, where digit recognition based on the MNIST data base is considered. In this highly dimensional
setting, specific scalability measures are required to apply our TADS technology.

\bigskip\noindent
After specifying the details of our running example in \cref{sec:XOR}, \cref{sec:ADS} sketches
Algebraic Decision Structures that later on will be instantiated with Affine Functions recalled in 
\cref{sec:PAF} to introduce the central notion of this paper, Typed Affine Decision Structures (TADS). Semantically, TADS represent piece-wise affine functions, which marks them as a 
fitting representation for Rectifier Networks that represent continuous piece-wise affine functions\footnote{Rectifier Networks are often also called Piece-wise Linear Neural Networks, the reason for us to denote them as PLNNs.} and that 
are discussed in \cref{sec:PLNN}.
Our main contribution is the derivation of TADS, using both symbolic execution and compositionality
along the layering structure of PLNN, as a complete and precise model explanation of PLNNs.
We introduce TADS in \cref{sec:tads} and state important algebraic properties that allow the manipulations mentioned beforehand.
Subsequently, \cref{sec:showcase} illustrates the impact on verification and explanation
of the algebraic properties of TADS that are also established in \cref{sec:tads} along the running example.
The paper closes after a discussion of related work in \cref{sec:related_work}
with conclusions and direction to future work in \cref{sec:conclusion}.

\section{Running Example -- XOR}%
\label{sec:XOR}%
\begin{figure}
\centering
\includegraphics[scale=.5]{"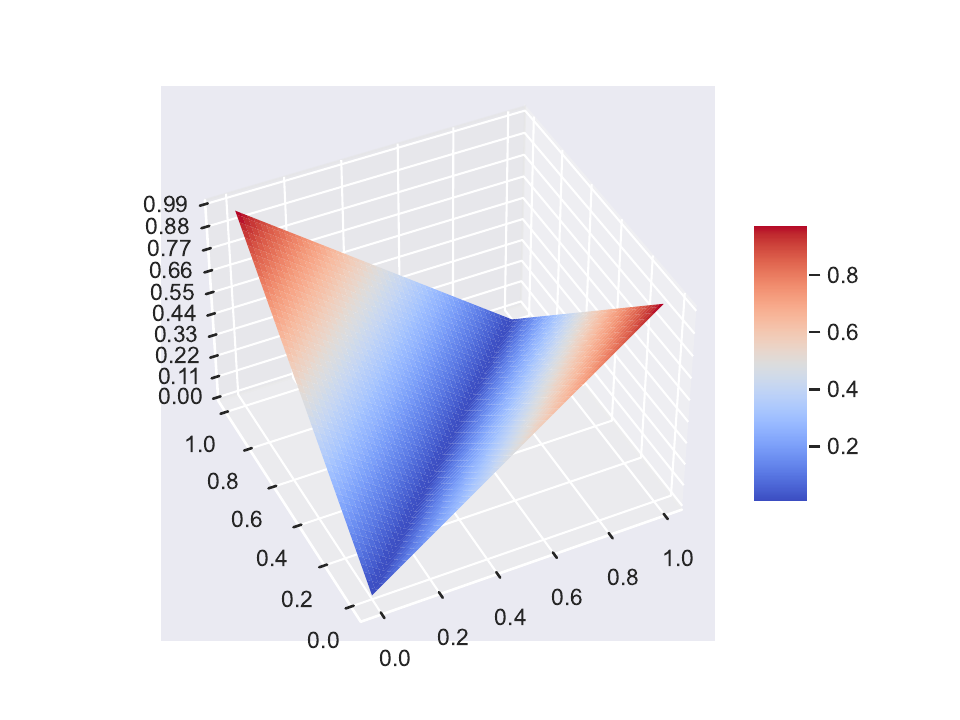"}
\caption{A baseline solution to the XOR-regression problem given by $f_*(x,y)= |x-y|$. Note that this function is piece-wise linear, having two separate linear regions, which is minimal for the problem.}
\label{fig:manualxor}
\end{figure}
\begin{figure}
\centering
\includegraphics[width=0.8\linewidth]{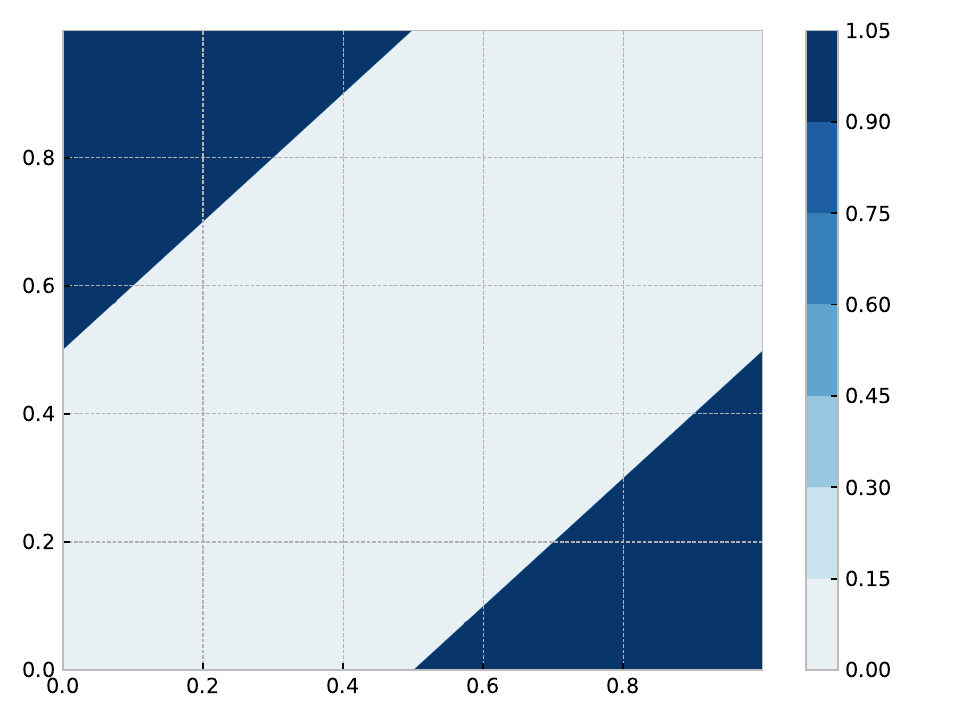}
\caption{A baseline solution to the XOR-classification problem given by $g_*(x,y)= 1$ iff $f_*(x,y) \geq 0.5$.}
\label{fig:manualxorclass}
\end{figure}
As a running example throughout this paper, we discuss the XOR function under the perspective of a 
regression task and a classification task, as specified below. We chose the XOR problem for
illustration for two reasons: 
\begin{itemize}
\item The XOR problem concerns a two-dimensional function which can be visualized as  a function plot.
\item While simple, the XOR problem has been a roadblock in early AI research because it cannot be solved by linear approaches \cite{minsky1969perceptrons}. Therefore, it is a minimal problem that still requires a more powerful non-linear model such as a PLNN.
\end{itemize}

\bigskip\noindent
For the following formalizations, let us fix some basic notation: The set of natural numbers (including zero) is denoted $\natset$ and the 
set of real numbers $\realset$.
Unions and intersections of sets are defined as usual.
The cartesian product of two sets is defined as 
\[ M \times N \defined \set{(x,y) \given x \in M \and y \in N } . \]
A sequence of cartesian products over the same set may be abbreviated as $M^n$ ($n \geq 0$), where
\begin{align*}
	M^0 &= \emptyset & 	M^1 &= M & M^{n+2} &= M \times M^{n+1}
\end{align*}
and the Kleene star operator is defined as
\[ M^* \defined \bigcup_{n\in\natset} M^n \]
Moreover, intervals of $\realset$ are of the form $[a,b]$ for $a, b \in \realset$ with $a \leq b$ and denote the set of all real values
between $a$ and~$b$.

\bigskip\noindent
As a \textit{regression} task, the XOR problem is stated as follows:
\begin{definition}[XOR Regression]
Find a piece-wise affine function $f\colon [0,1]^2 \rightarrow [0,1]$ that is continuous and satisfies:
\[f(0,0)\approx 0 \approx f(1,1) \quad \text{and} \quad f(1,0)\approx 1\approx f(0,1)\]
\end{definition}
Thus, a learning algorithm is tasked with approximating a continuous version of an XOR gate, interpolating between the four edge points for which the XOR function is defined.\footnote{The restriction to the interval $[0,1]$ is meant to ease the exposition.}

When posing the XOR problem as a classification task, the XOR function can be regarded as a function with \emph{discrete} binary output~$1$ or~$0$
but with a continuous domain $\realset^2$. 
\goodbreak
\begin{definition}[XOR Classification]
Find a piece-wise affine function $f\colon [0,1]^2 \rightarrow \{0,1 \}$ such that:
\[f(0,0)= 0=f(1,1) \quad \text{and} \quad f(1,0) =1= f(0,1)\]
\end{definition}
As the XOR-problem requires fixed values only at four points, there exist infinitely many solutions.  This is typical for machine learning problems where only some few points are fixed and others are left for the machine learning model to freely interpolate. Different machine learning models have different principles that dictate this interpolation.
For example, concerning PLNNs, the interpolation is (piece-wise) linear.

In line with the principle of Occam's razor~\cite{sober2015ockham}, humans\footnote{In contrast to, e.g., solutions learned by machines.} would optimally solve the XOR-regression problem with a function as simple as:
\[ f_*(x,y)= \abs{x-y} \]
A visualisation of $f_*$ can be found in \cref{fig:manualxor}.
Similarly, a human would probably choose the following  corresponding straightforward extension to the classification problem: 
\[g_*(x,y)= \begin{cases} 
	1 \quad &\text{if } f_*(x,y) \geq 0.5\\
	0 \quad &\text{otherwise}
\end{cases}\]
An illustration of $g_*$ can be found in \cref{fig:manualxorclass}.
It is straightforward to check that these functions solve the XOR-regression and XOR-classification problems optimally in the sense that they match the traditional XOR function at all points where it is defined. 

The continuous XOR problems will serve as running examples throughout this work:  We will demonstrate different representations of piece-wise linear functions (such as $f_*$) and transformations between them along the development of our theory, and showcase differences between the manually constructed solutions to the regression and classification tasks and their learned counterparts in \cref{sec:showcase}.

\section{Algebraic Decision Structures}%
\label{sec:ADS}%
In order to prepare the algebraic treatment of decision structures, we focus on decision structures whose leafs are labeled with the elements of an algebra $\algebraintro{A}{A}{O}$, so-called \emph{Algebraic Decision Structures} (ADDs). 
This subsumes the classical case where leafs of decision structures are elements of a set, as these are simply algebras where $\algoperations{O}$ is empty.
In this section we summarize the definitions and theorems of~\cite{gossen2021algebraic}, which are required later in this paper.\footnote{Some definitions and theorems were slightly improved or adjusted from~\cite{gossen2021algebraic} for better alignment with the rest of this paper. We omit the proofs for the adjustments because they are straightforward.}
\begin{definition}[Algebraic Structures]
	An \emph{algebraic structure}, or \emph{algebra} for short, is a pair $(\algcarrier{A},\algoperations{O})$ of a carrier set $\algcarrier{A}$ and a set of operations $\algoperations{O}$.
	Operations $\mathit{op} \in \algoperations{O}$ have a fixed arity and are closed under $\algcarrier{A}$.
\end{definition} 
In the following, the algebra is identified with its carrier set and both are written calligraphically.
\begin{definition}[Algebraic Decision Tree]
	\label{def:adt}%
	An Algebraic Decision Tree (ADT) over the algebra $\algebra{A}$ and the predicates $\mathcal{P}$\footnote{In contrast to ADDs, we do not require an ordering over $\mathcal{P}$ and therefore cannot guarantee canonicity.} is inductively defined by the following BNF:
	\[
	T \bnfeq \varepsilon \bnfmid a \bnfmid (p, T, T) \quad \textrm{ with } a \in \algcarrier{A} \textrm{ and } p \in \mathcal{P} \textrm{.}
	\]
	%
%	Let $\adtset{A}$ denote the set of all such ADTs.
\end{definition}
ADTs are (directed) trees where for a node $(p,l,r)$ the root is given by $p$ and the left and right children by $l$ and $r$, respectively.
ADTs of the form $a$ are leafs with no children and $\varepsilon$ denotes the empty tree.
We can merge nodes in these ADTs, which leads to the more general Algebraic Decision Structures~(ADS):
\begin{definition}[Node Merging]
	Let $t$ be an ADT and $t_1$ and $t_2$ be two nodes in $t$ such that $t_2$ is not reachable from $t_1$.
	Then, the two-step transformation of $t$
	\begin{itemize}
		\item re-route the incoming edges of $t_2$ to $t_1$ and
		\item eliminate all unreachable nodes of $t$.
	\end{itemize}
	is called a \emph{$t_2$ into $t_1$ merge}. 
\end{definition}
Node merges aggregate subtrees in a manner that does not create directed cycles.
Later, we will define semantics preserving merges.
\begin{definition}[Algebraic Decision Structure]
	\label{def:ads}%	
	A rooted directed acyclic graph (DAG) that results from an ADT by a series of node merges is called an Algebraic Decision Structure (ADS).
	Let $\adsset{A}$ denote the set of all such ADSs over an algebra $\algebra{A}$.
\end{definition}
We can define their semantics inductively.
\begin{definition}[Decision Structure Semantics]
	\label{def:ads_sem}%
	For a set $\Sigma$ of valuations the semantic function 
	\[ \adssem{} : \adsset{A} \rightarrow (\Sigma \rightarrow \algcarrier{A}) \]
	for ADSs is inductively defined as
	\begin{align*}
		%  	\adssem{\varepsilon}(\sigma) &\defined \varepsilon_A \\
		\adssem{a}(\sigma) &\defined a \\
		\adssem{(p, l, r)}(\sigma) &\defined
		\begin{cases}
			\adssem{l}(\sigma) &\textrm{if } \llbracket p \rrbracket(\sigma) = 1 \\
			\adssem{r}(\sigma) &\textrm{if } \llbracket p \rrbracket(\sigma) = 0 
		\end{cases}
	\end{align*}
\end{definition}
ADS can be considered the universe in which we operate, typically using semantics-preserving transformations.
In particular, we will frequently apply \emph{semantic reduction} and \emph{infeasible path reduction} as discussed in the next two subsections.
These two operations reduce the representational overhead of ADS while preserving their semantics.
\subsection{Semantic Reduction}%
\label{sec:sem-red}%
Semantic functions naturally induce an equivalence relation:
\begin{definition}[Semantic Equivalence]
	\label{def:sem_equivalence}%
	Two ADSs $t_1$ and $t_2$ are semantically equivalent iff their semantic functions coincide
	\[
	t_1 \sim t_2 \quad\textrm{iff}\quad \adssem{t_1} = \adssem{t_2}
	\]
\end{definition}
The following theorem states that one of two different nodes of an ADS that are semantically equivalent is redundant:
\begin{theorem}[Semantic Reduction]
	\label{theo:semanticReduction}%
	Let $t$ be an ADS with two nodes $t_1$ and $t_2$ that are semantically equivalent, i.e., $t_1 \sim t_2$, and such that $t_2$ is not reachable from $t_1$. 
	Moreover, let $t_3$ be the $t_2$ \emph{into} $t_1$ \emph{merge} of $t$. Then $t$ and $t_3$ are semantically equivalent, i.e., 
	$t \sim t_3$.
\end{theorem}
Our implementation heuristically reduces the number of semantically equivalent nodes of the ADSs. However, in contrast to Algebraic Decision Diagrams~\cite{bahar1997algebric},  which are known for their normal forms, we cannot guarantee canonicity here.

\subsection{Vacuity Reduction}%
\label{sec:unsat_paths}%
Typically, there are dependencies between different predicates in $\mathcal{P}$, which induces so-called infeasible paths in the corresponding ADSs.
This can be exploited for further reducing ADSs by eliminating so-called \emph{vacuous} predicates:
\newcommand{\final}{\mathit{final}}
\begin{definition}[Vacuity]
	\label{def:partial_vacuity}%
	Let $\bar{\mathcal{P}}$ be the set of negated predicates of $\mathcal{P}$. 
	Then we call  $\pi = p_0 \cdots p_{m} \in (\mathcal{P} \cup \bar{\mathcal{P}})^*$ a predicate path. 
	\begin{itemize}
		\item $\pi$ is called a predicate path of a decision structure $t \in \adsset{A}$ iff there exists a path $\pi' = p'_0 \cdots p'_{m} \in \mathcal{P}^*$ from the root of $t$ to one of its other nodes such that $p_i = p'_i$ in case that $\pi'$ follows the left/true branch at $p_i$ in $t$ and $p_i = \bar{p}'_i$ otherwise. 
		
		We denote the last predicate  $p_m \in \mathcal{P} \cup \bar{\mathcal{P}}$ of  $\pi$ by $\final(\pi)$.
		\item  Given a predicate path $\pi = p_0 \cdots p_{m}$ the predicate $\final(\pi)$
		is called \emph{vacuous} for $\pi$  iff
		the conjunction of the preceding predicates $p_0 \land \cdots \land p_{m-1}$ in $\pi$ implies $\final(\pi)$. 
		\item Let $\Pi_n$ be the set of predicate paths of $t \in \adsset{A}$ that end in a given node $n$.
		We call $n$ vacuous in $t$, iff $\final(\pi)$ is vacuous for all
		paths $\pi \in \Pi$ and $\final(\pi)$ coincides for all $\pi$.
		\item A decision structure $t \in \adsset{A}$ is called \emph{vacuity-free} iff there exists
		no vacuous node.
	\end{itemize}
\end{definition}
This allows us to define the following optimization step.
\begin{definition}[Vacuity Reduction] 
	\label{def:partial_vacuity_reduction}%
	Let $t \in \adsset{A}$ be a decision structure with a vacuous node $n$ and $\final(\pi) \in \mathcal{P} \cup \bar{\mathcal{P}}$ 
	be the last predicate of some predicate path $\pi$ ending in $n$. 
	Then, re-routing all incoming edges of $n$ to the 'true'-successor of $n$ in case of $\final(\pi) \in \mathcal{P}$  and to
	the 'false'-successor otherwise  is called a \emph{vacuity reduction} step. 
\end{definition}
ADSs, being DAGs, only have finitely many predicate paths which can be effectively analysed for vacuous predicates, as long as the individual predicates are decidable.
As the elimination of vacuous predicates is a simple semantics-preserving transformation, we have: 
\begin{theorem}[Minimality]
	\label{theo:minimality}%
	Every ADS can be effectively transformed into a semantically equivalent, vacuity-free ADS that is minimal in the sense that any further reduction would change its semantics.
\end{theorem}
In the remainder of the paper, we will not explicitly discuss the effects of semantic reduction and vacuity reduction. Rather, we will concentrate 
on the algebraic properties of ADS that they inherit from their leaf algebra via lifting.

\subsection{Lifting}%
\label{sec:lifting}%
It is well-known that algebraic structures $\algebraintro{A}{A}{O}$ can point-wise be lifted to cartesian products and arbitrary function spaces $M \to A$.
This has successfully been exploited for Binary Decision Diagrams (BDDs) and Algebraic Decision 
Diagrams (ADDs) that canonically represent functions of type $\binaryset^n \to \binaryset$ and $\binaryset^n \to \algcarrier{A}$ respectively.
In fact, the canonicity of these representations allows one to identify the BDD/ADD representations directly with their semantics, which in particular reduces the verification of semantic equivalence to checking for isomorphism.

\smallskip
\noindent
In our case, canonicity is unrealistic for two reasons (cf.,\ \cref{sec:tads-algebra}):
\begin{enumerate}
	\item Considering predicates rather than Boolean values 
	introduces infeasibility and thereby prohibits minimal
	canonical representations. 
	\item The ordering of predicates may lead to an exponential 
	explosion of the representation. Please note that, in contrast to, e.g.,  
	the typical BDD setting, we do  not have just a few (64, 128, 256,$\ldots$ or 
	the like) input bits that specify the control of some circuit, but predicates
	capture the effect of the ReLU function in a history-dependent way; Predicates that result from computations in a later layer depend on predicates from earlier layers. Moreover, as predicates are continuous objects, the probability of them coinciding can be considered $0$. Thus, ordering predicates would typically lead to representations that are doubly exponential in the number of neurons of a neural network.
\end{enumerate}
We will see, however, that all the algebraic properties we need also
hold for unordered ADSs, and that we can conveniently compute
on (arbitrary) representatives of the partition defined by semantic 
equivalence. This way, we arrive at an exponential worst-case complexity
(in the size of the argument PLNNs) both, for the algebraic operations 
and the decision of semantic equivalence. 

Although ADSs are not canonical one can effectively apply operators on concrete representatives while preserving semantics.
Every operator can be lifted inductively as follows
\begin{definition}[Lifted Operators]
	\label{def:lifted_operators}%
	For every operator $\genericoperator: \algcarrier{A}^2\to\algcarrier{A}$ of an algebra $\algebraintro{A}{A}{O}$ we define the lifted operator $\liftedgenericoperator\colon\adsset{A}{}^2\to\adsset{A}$ that operates over ADS  inductively as
	\begin{align*}
		a \liftedgenericoperator a' &= a \genericoperator a' \\
		a \liftedgenericoperator (p,l,r) &= (p, a \liftedgenericoperator l, a \liftedgenericoperator r) \\
		(p,l,r) \liftedgenericoperator t &= (p, l \liftedgenericoperator t, r \liftedgenericoperator t)
	\end{align*}
	where $a,a'\in \algcarrier{A}$ are ADS identified with an element of the algebra, $t,l,r\in \adsset{A}$ are ADS, and $p\in\mathcal{P}$ is a predicate.
\end{definition}
Intuitively, for two ADS $t_1$ and $t_2$, this construction replaces leaves in $t_1$ with copies of $t_2$. Thus, each path of the resulting ADS $t_3$ expresses a conjunction of one path in $t_1$ and one path in $t_2$. The partition of the domain imposed by all paths of  $t_2$ therefore coincides with the intersection imposed by the intersection of partitions imposed by $t_1$ and $t_2$.
The required lifting of the operators to leaf nodes is straightforward (cf.,\ \cref{fig:addition} for illustration).

The following theorem which states the correctness of the lifted operators can straightforwardly be proved by induction:
\begin{theorem}[Correctness of Lifted Operators]
	\label{theo:lifted_operators_correctness}%
	Let $t_1, t_2 \in \adsset{A}$ be two ADS over some algebra ${\algebraintro{A}{A}{O}}$.
	Let $\liftedgenericoperator\colon\adsset{A}{}^2\to\adsset{A}$ denote the lifted version of the operator $\genericoperator \in \algoperations{O}$.
	Then the following equation holds for all $\sigma\in\Sigma$:
	\[ \adssem{t_1 \liftedgenericoperator t_2}(\sigma) \defined \adssem{t_1}(\sigma) \genericoperator \adssem{t_2}(\sigma) \]
\end{theorem}
\subsection{Abstraction}
\label{sec:abstraction}%
Abstraction is one of the most powerful means for achieving scalability. The following
easy to prove theorem concerns the interplay of abstractions imposed by a homomorphism of the
leaf algebra and their effect on some classification function.
\begin{theorem}[Abstraction]
	\label{theo:lifting}%
	Let $\algebraintro{A}{A}{O}$ and $\algebraintro{A'}{A'}{O'}$ be two algebras, and
	$\alpha\colon \algebra{A} \rightarrow \algebra{A'}$ a homomorphism.
	Then  $\alpha_S\colon \adsset{A} \rightarrow \adsset{A'}$ defined by simply applying $\alpha$ to all the leaves of the argument ADS
	completes the following commutative diagram:
	\begin{center}
	\begin{tikzcd}
		\adsset{A} \ar[r, "\alpha_S"] \ar[d,"\adssem{}"] & \adsset{A'} \ar[d, "{\adssem[A']{}}"] \\
		\Sigma \to A \ar[r, "\alpha"] & \Sigma \to A'
	\end{tikzcd}
	\end{center}	
\end{theorem}
We will see in \cref{sec:showcase} how elegantly abstraction can be dealt with in
the TADS setting: The abstraction that transforms the XOR regression setting into a classification 
setting can be easily realized via the TADS composition operator.

\section{Affine Functions}
\label{sec:PAF}%
The following notations of linear algebra are based on the book~\cite{axler1997linear}.
The real vector space $(\realset^n, +, \cdot)$ with $n > 0$ is an algebraic structure with the operations
\begin{align*}
	+ &\colon \realset^n \times \realset^n \to \realset^n & &\text{vector addition}\\
	{}\cdot{} &\colon \realset\phantom{{}^n} \times \realset^n \to \realset^n & &\text{scalar multiplication}
\end{align*}
which are defined as
\begin{align*}
	(x_1,\dots, x_n) + (y_1,\dots,y_n) &= (x_1 + y_1, \dots, x_n + y_n) \\
	\lambda \cdot (x_1,\dots,x_n) &= (\lambda \cdot x_1, \dots, \lambda \cdot x_n)
\end{align*}
A real vector $(x_1,\dots,x_n)$ of $\realset^n$ is abbreviated as $\vec x$.
To refer to one of the components, we write $\vec{x}_i \defined x_i$ (note the arrow ends before the subscript in contrast to $\vec{x_i}$, which denotes the $i$-th vector).
The dimension of a real vector space $\realset^n$ is given as $\dim \realset^n = n$.

A matrix $\mat W$ is a collection of real values arranged in a rectangular array with $n$ rows and $m$ columns.
\[ \mat W = 
\begin{pmatrix}
	w_{1,1} &w_{1,2} &\dots  &w_{1,m} \\
	w_{2,1} &w_{2,2} &\dots  &w_{2,m} \\
	\vdots  &\vdots  &\ddots &\vdots \\
	w_{n,1} &w_{n,2} &\dots  &w_{n,m}
\end{pmatrix} \]
To indicate the number of rows and columns, one says $\mat W$ has \emph{type} $n \times m$ also notated as $\mat W \in \realset^{n \times m}$.

An element at position $i,j$ of the matrix $\mat{W}$ is denoted by $\mat{W}_{i,j} \defined w_{i,j}$ (where $1 \leq i \leq n$ and $1 \leq j \leq m$).
A matrix $\mat W \in \realset^{n\times m}$ can be reflected along the main diagonal resulting in the transpose $\mat{W}^\transp$ of shape $m \times n$ defined by the equation
\[ \big(\: \mat{W}^\transp \:\big)_{i,j} \defined \mat{W}_{j,i} \]
The $i$-th row of $\mat W$ can be regarded as a $1 \times m$ matrix given by
\[ \mat{W}_{i,\cdot} \defined (w_{i,1}, \dots, w_{i,m}) .\]
Similarly, the $j$-th column of $\mat W$ can be regarded as a $n \times 1$ matrix defined as
\[\mat{W}_{\cdot,j} \defined (w_{1,j}, \dots, w_{n,j})^\transp .\]
Matrix addition is defined over matrices with the same type to be component-wise, i.e., 
\[ \big(\: \mat{W} + \mat{N} \:\big)_{i,j} \defined \mat{W}_{i,j} + \mat{N}_{i,j} \]
and scalar multiplication as
\[ \big(\: \lambda \cdot \mat{W} \:\big)_{i,j} \defined \lambda \cdot \mat{W}_{i,j} . \]
The (type-correct) multiplication of two matrices ${\mat W \in \realset^{n\times r}}$ and $\mat N \in \realset^{r\times m}$ is defined as
\[ \big(\: \mat W \cdot \mat N \:\big)_{i,j} \defined \sum_{k=1}^r \mat{W}_{i,k}\cdot \mat{N}_{k,j} \]
Identifying 
\begin{itemize}
	\item $n \times 1$ matrices with (column) vectors
	\item $1 \times m$ matrices with row vectors
	\item $1 \times 1$ matrices with scalars
\end{itemize}
as indicated above, makes the well-known dot product of $\vec v, \vec w \in \realset^n$
\[ \dotprod{\vec v}{\vec w} \defined \sum_{i=1}^{n} \vec{v}_i \cdot \vec{w}_i \]
just a special case of matrix multiplication.
The same holds for matrix-vector multiplication that is defined for a $n \times m$ matrix $\mat W$ and a vector $\vec x \in \realset^n$ as
\[ \big(\: \mat W \cdot \vec x \:\big)_{i} \defined \mat W_{i,\cdot} \cdot \vec x \]
Matrices with the same number of rows and columns, i.e., with type $n\times n$ for some $n\in\natset$, are said to be \emph{square matrices}.
Square matrices have a neutral element for matrix multiplication, called \emph{identity matrix}, that is zero everywhere except for the entries on the main diagonal which are one.
\[ \idmat{n} \defined 
\begin{pmatrix}
	1 & &0 \\
	 &\ddots &\\
	0  &  &1 \\
\end{pmatrix} \]
%
%A very similar matrix is the defect matrix $\idzmat{n}{i}$ which is like the identity matrix but the $(i,i)$-th entry on the main diagonal is also zero.
The $i$-th unit vector $\unitvec{i}$ is a vector where all entries are zero except the $i$-th which is one.

\subsection{Piece-wise Affine Functions}
\begin{definition}[Affine Function]
	\label{def:affine_functions}%
	A function $\aff\colon \realset^{n} \rightarrow \realset^m$ is called \emph{affine} iff it can be written as 
	\[ \aff(\vec x)= \mat{W} \vec x + \vec b \]
	for some matrix $\mat W \in \realset^{m \times n}$ and vector $b \in \realset^m$.\footnote{In the context of neural networks, the weights $\mat W$ and bias $\vec b$ are the result of some learning procedure. In this work, we assume that they are always known and fixed.}

	We identify the semantics and syntax of affine functions with the pair $(\mat W, \vec b)$, which can be considered 
	as a canonical representation of affine functions.
	Furthermore, we denote the set of all affine functions $\realset^n \rightarrow \realset^m$ as $\affset(n,m)$
	and call $(n,m)$ the \emph{type} of $\affset(n,m)$.
	The untyped version $\affset$ is meant to refer to the set of all affine functions, independently of their type.
\end{definition}
\begin{lemma}[Operations on Affine Functions]%
	\label{theo:affine_operations}%
	Let $\aff_1,\aff_2$ be two affine functions in canonical form, i.e., 
	\begin{align*}
		\aff_1(\vec x) &= \mat{W_1} \vec x + \vec{b_1} \\
		\aff_2(\vec x) &= \mat{W_2} \vec x + \vec{b_2}
	\end{align*}
	Assuming matching types, the operations $+$ (addition), $\cdot$ (scalar multiplication), and $\circ$ (function application) can be calculated on the representation as
	\begin{align*}
		(s \cdot \aff_1)(\vec x)
		&= (s \cdot \mat {W_1}) \,\vec x + (s \cdot \vec {b_1}) \\
		(\aff_1 + \aff_2)(\vec x)
		&= (\mat {W_1}+\mat {W_2}) \,\vec x + (\vec {b_1} + \vec {b_2}) \\
		(\aff_2 \circ \aff_1)(\vec x)
		&= (\mat {W_2} \mat {W_1}) \,\vec x + (\mat {W_2} \vec {b_1} + \vec {b_2})
	\end{align*}
	resulting again in an affine function in canonical representation.
\end{lemma}
It is well-known that the type resulting from function composition evolves as follows
\[ \circ\colon \affset(r,m) \times \affset(n, r) \to \affset(n,m) . \]
The type of the operation is important for the closure axiom, the basis for most algebraic structures. 
This leads to the following well-known theorem~\cite{axler1997linear}:
\begin{theorem}[Algebraic Properties]
	\label{prop:algprop}%
	Denoting, as usual, scalar multiplication with $\cdot$ and
	function composition with $\circ$, we have:
	\begin{itemize}[label={--},noitemsep]
		\item
		$(\affset(n,m), +, \cdot) $ forms a vector space and
		\item 
		$(\affset(n,n), \circ) $ forms a monoid. 
	\end{itemize}
\end{theorem}
This theorem can straightforwardly be lifted to untyped $\affset$ by simply restricting all operations 
to the cases where they are well-typed, i.e., where addition  is restricted to functions of the same type ($+_t$), and 
function composition to situation where the output type of the first function matches the input type of the second ($\circ_t$): 

\begin{theorem}[Properties of Typed Operations]%
	\label{prop:typed-algprop}%
	The quadruple $(\affset, +_t, \cdot, \circ_t) $ forms a typed algebra, i.e., an algebraic structure that is closed under well-typed operations.
\end{theorem}
Piece-wise affine functions are usually defined over a polyhedral partitioning of the pre-image space~\cite{Brondsted93,GorokhovikZ94,Ovchinnikov2010}.
\begin{definition}[Halfspace]
	Let $\vec w \in \realset^n$ and $b \in \realset$.
	Then the set 
	\[ p = \set{\vec x \in \realset^n \given \dotprod{\vec w}{\vec x} + b = 0} \]
	is called a \emph{hyperplane} of $\realset^n$.
	A hyperplane partitions $\realset^n$ into two convex subspaces, called \emph{halfspaces}.
	The positive and negative halfspaces of $p$ are defined as
	\begin{align*}
		p^+ &\defined \set{\vec x \in \realset^n \given \dotprod{\vec w}{\vec x} + b > 0 } \\
		p^- &\defined \set{\vec x \in \realset^n \given \dotprod{\vec w}{\vec x} + b < 0 }
	\end{align*}
	%The set of all halfspaces for some 
	%$\vec w \in \realset^d$ and $b \in \realset$
	%$$\set{ \set{x \in \realset^d \given \dotprod{\vec w}{\vec x} + b \leq 0} \given \vec w \in \realset^d, b \in \realset }$$
	%is denoted by  $\HS_d$.
\end{definition}
\begin{definition}[Polyhedron]     
	A polyhedron $Q \subseteq \realset^n$ is the intersection of $k$ halfspaces for some
	natural number $k$.
	\[ Q=\bigcap_{i=1}^k \set{ \vec x \in \realset^n \given \dotprod{\vec {w_i}}{\vec x} + {b_i} \leq 0 } \]
\end{definition}

\begin{definition}[Piece-wise Affine Function]
	A function $\paff\colon \realset^{n} \rightarrow \realset^m$ is called \textit{piece-wise affine} if it can be written as
	$$	\paff(\vec x)= \aff_i(\vec x) \; \text{ for } \; \vec x \in Q_i  $$
	where $Q = \set{Q_1,\dots,Q_k}$ is a set of polyhedra that partitions the space of $\vec x$ and $\aff_1, \dots, \aff_k$ are some affine functions. We call $\aff_i=\mat W_i \vec x +\vec b_i$ ($1\leq i\leq k$) the function associated with polyhedron $Q_i$.
	%We denote the set of piece-wisse affine functions by $\pwaset$.
\end{definition}
The proof of the following property is straightforward: 
\begin{proposition}[Continuity]
\label{prop:continuity}%
A piece-wise affine function is continuous iff at each border between two connected polyhedra the affine functions associated with either polygon agree.
\end{proposition}

\subsection{The Activation Function ReLU}
\label{sec:ReLU}%
In this paper, we focus on neural network architectures that use the ReLU activation function:
\begin{definition}[ReLU]
	The Rectified Linear Unit (ReLU) 
	\[ \Relu^k : \realset^k \rightarrow \realset_+^k \]
	 is a projection of $\realset^k$ onto the space of positive vectors $\realset_+^k$
	defined by replacing each component $x_i$ of a vector $\vec x$ by  $\max\set{0,x_i}$:
	\[ \big(\,\Relu^k(\vec x) \,\big)_j \defined 
		\max \set{0, x_j} 
	\]
	If the input dimension is clear, we omit the index and just write $\Relu$.
\end{definition}

The term $\max \set{0,x_i}$ is continuous and piece-wise linear. As $\Relu$ operates independently on all dimensions of its input, it is itself piece-wise linear.

From a practical perspective, $\Relu$ is one of the  best understood activation functions, and the corresponding
rectifier networks are one of the most popular modern neural network architectures~\cite{glorot2011deep}.

For ease of notation in later sections, we use the fact that $\Relu$ operates on each component of a vector independently, and can therefore be decomposed into
\begin{equation*}
	\Relu^k = \relu{k}{k} \circ \relu{k}{k-1} \circ \dots \circ \relu{k}{1}
\end{equation*}
where $\relu{k}{i} : \realset^k \rightarrow \realset^k$ is the \emph{partial ReLU function} defined by setting the $i$-th component of a vector $\vec x$ to $0$ iff $x_i < 0$.
More formally,
\[ \big(\,\relu{k}{i}(\vec x) \,\big)_j \defined 
\begin{cases}
	x_j \quad &\text{if } i \neq j \\
	\max \set{0, x_j} &\text{if } i = j
\end{cases} \:.\]
%\smallskip
 
\section{Piece-wise Linear Neural Network}
\label{sec:PLNN}%
Piece-wise linear neural networks are specific representations of continuous piece-wise affine functions.
Calling them piece-wise linear is formally incorrect  (the term piece-wise affine would be correct), but established. 
For the ease of exposition, we restrict the following development to the case where all activation functions are partial ReLU functions. This suffices to capture the entire class of Rectifier Networks, which themselves can represent all piece-wise affine functions~\cite{he2018relu}.
We adopt the popular naming in the following definition:
\begin{definition}[Rectifier Neural Networks]
	\label{def:PLNN}%
	The syntax for \emph{Rectifier Neural Networks}, or here synonymously used, \emph{Piece-wise Linear Neural Networks}
	(PLNNs),  is defined by the following BNF
	\[
		\plnnset \bnfeq \varepsilon \bnfmid \alpha \netcomp {\plnnset} \bnfmid \phi \netcomp \plnnset 
	\]
	where the meta variables $\alpha$ and $\phi$ stand for affine functions and partial ReLU functions, respectively. 
	Writing PLNNs as $N = f_0\netcomp\dots\netcomp f_l$ where $f\in\set{\alpha,\phi}$
	we denote the set of all PLNNs with $\domain(f_0)= \realset^{n}$ and $\codomain(f_{l})= \realset^{m}$ as $\plnnset(n, m)$ and the set of all PLNNs as 
	\[
	\plnnset = \bigcup_{ n, m \in \natset}\plnnset(n, m)
	\]
\end{definition} 
This definition of a PLNN slightly flexibilizes the classical definition as it does not require the strict 
alternation of affine functions and activation functions and uses partial ReLU functions instead of ReLU\@. We will exploit this flexibility to directly have the right granularity for defining according operational semantics (cf.,\ \cref{sec:sos}). 

\begin{example}[Representing XOR as PLNN]
	As stated in \cref{sec:XOR}, our baseline solution to the XOR regression model is defined by the function $\abs{x-y}$.
	We can represent this function as a PLNN $N_{*}$.
	It consists of two affine functions
	\begin{align*}
		\alpha_1 &=
		\begin{pmatrix}
			1 & -1 \\
			-1 & 1
		\end{pmatrix} &
		\alpha_2 &=
		\begin{pmatrix}
			1 & 1
		\end{pmatrix}
	\end{align*}
	and two partial ReLUs applied in this order:
	\[ N_{*} \defined \alpha_1\netcomp\relu{2}{1}\netcomp\relu{2}{2}\netcomp\alpha_2 \] 
	Note that typically $N_*$ would be defined as 
	\[ N_*' = \alpha_1\netcomp{\Relu}\netcomp\alpha_2 \]
	 in the context of machine learning.
	However, as both definitions impose the same semantics 
	\[ \netsem{N_*} = \netsem{N_*'}\]
	we defined it directly using the notational conventions of this paper.
	This construction uses the observation that 
	\[\Relu(x-y) = \begin{cases}
		x-y \quad &\text{if } x>y \\
		0 &\text{otherwise}
	\end{cases} \ .\]
	The following figure shows the (instantiated) corresponding
	network architecture:
	\begin{center}
		\includegraphics[width=\columnwidth,height=75pt,keepaspectratio]{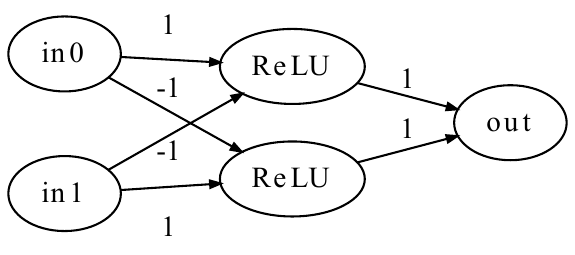}
	\end{center}
	This example shows that PLNNs can encode our baseline solution to the XOR problem. However, it is important to note that PLNNs are not usually manually defined, but rather trained to approximate a function using approximate learning algorithms~\cite{Goodfellow-et-al-2016}, see also \cref{sec:showcase}.
\end{example}

\subsection{Semantics of PLNNs}
PLNNs come with a very natural denotational semantics:
\begin{definition}[Denotational Semantics]%
	\label{def:network_denotational}%
	The denotational semantics 
	\[\semds{} : \plnnset(n, m) \rightarrow (\realset^{n} \rightarrow \realset^{m})\] 
	of PLNNs is inductively defined as the composition of all functions in a PLNN:
	\[
	\semds{\varepsilon} = \idfunc \ \ \text{and} \ \ \semds{f\netcomp N} = \semds{N} \circ f
	\]
	where $f \in \{\alpha, \phi\}$.
\end{definition}
Remark: In this definition we overload $f$ to represent both its corresponding syntactic artifact (e.g., a matrix) and its semantic artifact (e.g., the corresponding affine function or partial ReLU function, respectively). In the remainder of the paper it should always be clear from the context which interpretation we refer to.

PLNNs can also be evaluated in an operational manner based on derivation rules which closely resemble their process of computation.
For that we define the defect matrix $\idzmat{k}{i}$ ($1\leq i\leq k$) as the identity matrix $\idmat{k} $ where the $i$-th entry on the main diagonal, i.e., element $(i,i)$, is set to zero.
\begin{definition}[Operational Semantics]%
	\label{def:network_operational_rules}%
	The operational semantics of PLNNs is defined via the following three rules that operate on configurations ${\config{N, \vec x}}$ consisting of the remainder $N$ of the PLNN to process and the current intermediate result vector $\vec x$.
	\begin{align*}
		&\inference[Affine]{\text{$\alpha$ is affine}}{\config{\alpha\netcomp N, \vec x} \sosto{\true}  \config{N, \alpha(\vec x)}}\\[1mm]
		&\inference[ReLU 1]{x_i \geq 0}{\config{ \relu{k}{i} \netcomp N , \vec x} \sosto{1}  \config{N, \vec x}} \\[1mm]
		&\inference[ReLU 2]{x_i <0}{\config{\relu{k}{i} \netcomp N , \vec x}\sosto{0}  \config{N,  \idzmat{k}{i} \vec x}}	
	\end{align*}
\end{definition}
The labels (the symbols above the arrow) provide a history of which rule was applied.
It is easy to see that the rule to be applied next is always uniquely determined by the first component (the PLNN) which guarantees that the operational semantics is deterministic.
In fact, for each input there exists a unique computation path. Thus, the following definition is well-defined:

\begin{definition}[Semantic Functional $\semos{}$]%
\label{def:network_operational}%
	The semantic functional for the operational semantics 
	\[\semos{} : \plnnset(n, m) \rightarrow (\realset^{n} \rightarrow \realset^{m})\] 
	is defined as
	\[\semos{N}(\vec x)= \vec y \ \  \mbox{iff}\ \  \config{N, \vec x} \sosto{}^* \config{\varepsilon, \vec{y}}\]
\end{definition}
Note that these rules stand in close correspondence to the denotational semantics of PLNNs with each rule describing the evaluation of one of its constituent functions. 
In fact, we have:
\begin{theorem}[Correctness of $\semos{}$]%
	For any $N \in \plnnset$ we have $\semds{N} = \semos{N}$.
\end{theorem}
\begin{sketch}
The proof follows straightforwardly by induction on the number of layers of a PLNN\@. It suffices to show that the affine rule corresponds to the application of the affine function $\alpha$ and that the executions of the adequate rules ReLU~1 and ReLU~2 correctly cover the partial ReLU activation functions.
\end{sketch}
Thus, the operational semantics $\semos{}$ provides a constructive, local, and correct semantic interpretation of PLNNs.
\begin{example}[Semantics of $N_{*}$]%
	We consider the baseline solution to the XOR regression model  defined by the function $\abs{x-y}$.
	The network $N_{*}$ implements this function as a PLNN\@.
	We calculate $\semos{N_{*}}$ by applying the SOS rules to the initial configuration $\config{N_{*},(1,0)^\transp}$.
	\begin{align*}
		&\config{\alpha_1\netcomp\relu{2}{1}\netcomp\relu{2}{2}\netcomp\alpha_2, 
		(1, 0)^\transp} \\
		\sosto{\true} \quad &\config{\relu{2}{1}\netcomp\relu{2}{2}\netcomp\alpha_2, 
			(1, -1)^\transp} \\
		\sosto{1} \quad &\config{\relu{2}{2}\netcomp\alpha_2, 
			(1, -1)^\transp} \\
		\sosto{0} \quad &\config{\alpha_2, 
			(1, 0)^\transp} \\
		\sosto{\true} \quad &\config{\varepsilon, 
			1}
	\end{align*}
	This is the correct outcome. Note that the SOS rules correspond to an iterative processing of each component function (i.e., layer) of the neural network, much like function composition.
\end{example}
Next, we will naturally adapt the presented rules to symbolic execution, which by itself provides the first outcome explanation of PLNNs.

\subsection{Symbolic Execution of PLNNs}
\label{sec:sos}%
Symbolic execution aims at characterizing program 
states in terms of 
symbolic input values and corresponding path conditions. In particular,
it reveals how program states depend on the initial values during execution.
PLNNs are ideally suited for symbolic execution as they are acyclic computation graphs and contain only affine computations.
\begin{itemize}
	\item Affine functions are closed under composition. This allows one to aggregate (partially evaluate) the entire symbolic computation 
	history corresponding to some symbolic execution path in terms of a single affine 
	function, and to express all paths conditions as affine inequalities, also expressed in 
	terms of the initial values. 
	\item PLNNs possess finite, acyclic computation graphs, which conceptually allow for precise execution without need for abstractions.
\end{itemize}
In \cref{sec:tads}, we will see that this results in a directed, acyclic, side-effect-free computation graph whose leaves are affine function in $\affset(\inn,\out)$
that express the PLNN's effect on inputs belonging to the polyhedron specified by the path condition.

We define the required symbolic execution via derivation rules that transform configurations of the form $\config{N, \aff, \pi}$, where
\begin{itemize}
	\item $N \in \plnnset(r, m)$,
	\item $\aff\colon\realset^n\to\realset^r$ with representation $\aff(\vec x)=\mat W\vec x + \vec b$,
	\item and $\pi$ is a path condition
\end{itemize}
throughout the transformation.
The dimensions of $n$ and $m$ are bound by the initial PLNN while $r$ is the dimension of some hidden layer.
The following definition operates on the concrete representations of $N$, $\aff$, and $\pc$.
In the case of the last two, the representation is expected to be canonical and therefore syntax and semantics can be identified.\footnote{By using canonical representations it is impossible to trace the history of operations. One effectively cannot distinguish between isomorphic objects.}
Operations are expected to be applied directly to the representation.
Thus, the effect of a concrete execution path of $\semos{}$ is \emph{aggregated} (instead of simply recorded) into the components $\aff$ and $\pi$, while $N$ is destructed further and further until all layers have been considered.
\begin{definition}[Symbolic Execution of PLNNs]
\begin{align*}
    \config{ \alpha'\netcomp N, \aff,\pc} &\symderiv{\true}  \config{N, {\alpha' \circ \aff},\pc} \\[1mm]
    \config{ \relu{k}{i} \netcomp N, \aff,\pc}&\symderiv{1}  \config{ N,  \aff,{\pc' \land \pc}} \\[1mm]
    \config{ \relu{k}{i} \netcomp N, \aff,\pc}&\symderiv{0}  \config{ N, {\idzmat{k}{i}\circ \aff},{\neg \pc' \land \pc}}
\end{align*}
where $\pi ' = \aff(x)_i \geq 0$. 
\end{definition}
For a sequence 
\[ c_0 \symderiv{a_1} c_1 \:\cdots\: c_{n-1} \symderiv{a_n} c_n \]
of derivations we write $c_0 \symderiv{a_1\cdots a_n} c_n$.
Further, we denote with $(\symderiv{})^k$ the application of $\symderiv{}$ $k$ times, and
we write $\symderiv{}^*$ if $k$ is of no interest.
The following properties follow by straightforward induction on the length of the derivation sequences:
\begin{proposition}[Derivation Sequences]
	The following properties hold for all derivations of $\symderiv{}$:
	\begin{enumerate}
		\item $\displaystyle \config{N,\idfunc,\true} \symderiv{w} \config{\varepsilon, \aff, \pc} \iff$ \\
		$\displaystyle \config{N\netcomp N',\idfunc,\true} \symderiv{w} \config{N', \aff, \pc}$,
		\item $\displaystyle \config{N,\idfunc,\true} \symderiv{w} \config{\varepsilon, \aff, \pc} \implies$ \\
		$\displaystyle \config{N,\aff',\pc'} \symderiv{w} \config{\varepsilon, \aff \circ \aff', \pc \land \pc'}$,
		\item $\displaystyle \config{N,\aff,\pc} \symderiv{w} \config{N', \aff', \pc'}$ is unique in $w$.
	\end{enumerate}
\end{proposition}
Intuitively, the first identity states that derivations with the same prefix in the first component result in the same configuration after the prefix was completely processed.
The second states the effect of other starting values in the initial configuration.
Note that this relation does not hold in the reversed case, as $\circ$ and $\land$ are not injective and the configuration is only determined up to isomorphism.
The last identity corresponds to the result that $\sosto{}$ of \cref{def:network_operational_rules} is uniquely determined.
As the symbolic rules are more general, the result is restricted to the case where the word $w$ is known.

Moreover, the path conditions induce a partition of the input space $\realset^n$.
\begin{lemma}[Partition of $\pc$]
	\label{theo:path_condition_partition}%
	For an arbitrary but fixed $N\in\plnnset$ define the set of all derivations with depth $k$ as
	\[ V_k(N) \defined \set{ c \given \config{N,\idfunc,\true} \symderiv{w} c \: \land \: \abs{w} = k }  \]
	Define the set of all path conditions of the same $V_k$ as $\Pi_k$, then
	\begin{itemize}
		\item each $\pi\in\Pi_k$ defines a polyhedron for $k>0$  
		\item the polyhedra of $\Pi_k$ are a partition of $\realset^n$.
	\end{itemize}
\end{lemma}
\begin{sketch}
	Induction over derivation sequences of $N$.
\end{sketch}
Specifically, for each input vector $\vec x \in \realset^n$, there exists exactly one sequence of derivations $\displaystyle \config{N,\idfunc,\true} \symderiv{w} \config{\varepsilon, \aff, \pc}$ 
such that $\pi(\vec x)$ holds.
Therefore, the following is well-defined:
\begin{definition}[Semantic Functional $\ssemos{}$]
	The semantic functional for the symbolic operational semantics 
	$$\ssemos{} : \plnnset(\inn, \out) \rightarrow (\realset^{\inn} \rightarrow \realset^{\out})$$   is defined as
	 $\ssemos{N}(\vec x)= \vec y$   iff
	$$
	  \config{N, \idfunc, \true} \symderiv{}^*  \config{\varepsilon, \aff, \pi}  \ \ \land \ \ \pc(\vec x) = 1 \ \ \land \ \ \aff(\vec x) = \vec y
	$$
\end{definition}
Also the symbolic operational semantics is fully aligned with the denotational semantics:

\begin{theorem}[Correctness of $\ssemos{}$]
	\label{theo:semos}%
	For any $N \in \plnnset$ we have:  \ \ $\semds{N} = \ssemos{N}$.
\end{theorem}
\begin{sketch}
According to \cref{theo:semos}, it suffices to
show the semantic equivalence with $\semos{}$. As both the concrete and the symbolic operational semantics define unique computation paths for each input vector, the proof follows straightforwardly by an inductive proof that 
establishes the desired equivalence as an invariant when simultaneously following these paths. More concretely, we can prove
$$\forall \vec x \in \realset^{\inn} : \semos{N}(\vec x) = \ssemos{N}(\vec x)$$ using the following
induction hypothesis
\begin{align*}
	&\config{N_0, \vec{x_0}} \sosto{w}_{\scriptscriptstyle \mathrm{OS}} \config{N_k, \vec{x_k}} \iff \\
	&\config{N_0, \idfunc, \true} \symderiv{w} \config{N_k, \aff_k, \pc_k} \land \alpha_k(\vec {x_0}) = \vec{x_k} \land \pc_k(\vec {x_0})
\end{align*}
by a simple analysis of the following three cases: 
\begin{enumerate}
	\item $\displaystyle N_{k+1} = \alpha'\netcomp N_{k}$
	\item $\displaystyle N_{k+1} = \relu{k}{i}\netcomp N_{k} \ \land \ \vec{x}_i \geq 0$
	\item $\displaystyle N_{k+1} = \relu{k}{i}\netcomp N_{k} \ \land \ \vec{x}_i < 0$
\end{enumerate}
\end{sketch}
The symbolic operational semantics of PLNNs is sufficient to derive local explanations and decision boundaries similar to the ones presented in~\cite{chu2018exact,gopinath2018symbolic}.
In the following, we will show how symbolic operational semantics can be used to define semantically equivalent
Typed Affine Decisions Structures (TADS), which themselves are specific Algebraic Decision Structures (ADS), as defined in the
next section.
TADS collect all the local explanations in an efficient query structure such that we arrive at model explanations.

\begin{example}[XOR-Regression]
	As a brief example for the symbolic execution of PLNNs, we will calculate $\ssemos{N_{*}}$ by applying the symbolic SOS rules to the initial configuration $\config{N_{*},\idfunc, \true}$.
	Symbolic interpretation is not deterministic for the partial ReLU functions. We therefore chose the execution path that corresponds to the former example $\vec x = (1,0)^\transp$, i.e., with the label sequence $w=(\true,1,0,\true)$, for illustration:
	\begin{align*}
		&\config{\alpha_1\netcomp\relu{2}{1}\netcomp\relu{2}{2}\netcomp\alpha_2, 
			\begin{pmatrix}
				1 & 0\\
				0 & 1
			\end{pmatrix},
			\true} \\
		\symderiv{\true} \quad &\config{\relu{2}{1}\netcomp\relu{2}{2}\netcomp\alpha_2, 
			\begin{pmatrix}
				1 & -1\\
				-1 & 1
			\end{pmatrix},
			\true} \\
		\symderiv{1} \quad &\config{\relu{2}{2}\netcomp\alpha_2, 
			\begin{pmatrix}
				1 & -1\\
				-1 & 1
			\end{pmatrix},
			x_1 - x_2 \geq 0} \\
		\symderiv{0} \quad &\config{\alpha_2, 
			\begin{pmatrix}
				1 & -1\\
				0 & 0
			\end{pmatrix},
			x_1 - x_2 > 0} \\
		\symderiv{\true} \quad &\config{\varepsilon, 
			\begin{pmatrix}
				1 & -1
			\end{pmatrix},
			x_1 - x_2 > 0}
	\end{align*}
	Note that the path conditions and the affine functions have been simplified in every step.
	The affine functions are given in their canonical representation $\mat W \vec x + \vec b$ (as $\vec b$ is zero in all steps it is omitted).
	For the path conditions we have not fixed a representation, instead they are simplified to aid readability.
	The most important simplifications are
	\begin{align*}
		\left(
		\begin{pmatrix}
			1 & -1\\
			-1 & 1
		\end{pmatrix} \vec x \right)_1 \geq 0 &\iff x_1 - x_2 \geq 0 \\
		\lnot (-x_1 + x_2 \geq 0) \land x_1 - x_2 \geq 0 &\iff x_1 - x_2 > 0
	\end{align*}
\end{example}

\section{Typed Affine Decision Structures}
\label{sec:tads}
Consider the transition system $(V, \symderiv{})$ that represents the symbolic operational semantics $\ssemos{}$ of some ${N \in \plnnset(n,m)}$ where
\[ V = \set{c \given \config{N,\idfunc,\true}\symderiv{}^* c } \]
is the set of configurations which are reachable from  $\config{N, \mathrm{id}, \true}$
and let (recall \cref{def:ads,def:affine_functions})
\[ \tau \colon V \to \adsset{\affset} \]
denote the following inductively defined transformation that closely follows the symbolic SOS rules:

\begin{itemize}
	\item $\tau(\config{ \varepsilon, \aff, {}\cdot{}})   \  \defined \ \aff  $
	\item  $\tau(\config{ \aff' \netcomp N, \aff, {}\cdot{}}) \ \defined \  (\true, \tau(\config{ N, \aff' \circ \aff, {}\cdot{}}), \varepsilon)$
	\item $\tau(\config{ \relu{k}{i} \netcomp N, \aff, {}\cdot{}})  \\ \defined \ (\aff(x)_i \geq 0,  \tau(\config{N, \aff, {}\cdot{}}), \tau(\config{N, \idzmat{k}{i} \circ \aff, {}\cdot{}) })$
\end{itemize}
where \enquote{${}\cdot{}$} should be considered a don't care entry. Identifying $N$ with its computation tree, which is specified by its set of configurations that are
reachable from  $\config{N, \mathrm{id}, \true}$,\footnote{Please note that the transition labels $\true, 1,$ and $0$ are redundant.}  $\tau$ can be regarded 
as an injective relabeling of this tree, which results in the structure of an ADT:
\begin{theorem}[TADT]
	\label{def:TADT}%
	Let $N \in \plnnset(n,m)$. Then $\tau(N)$ is an ADT over $\affset(n,m)$ whose predicates are all of the form of affine inequalities.
\end{theorem}
\begin{sketch}
	The proof follows by straightforward induction along the isomorphic structure of the two trees.
	The following invariants hold for all steps of the transformation
	\begin{align*}
		\tau(c) = (p,\tau(c_t),{}\cdot{}) &\iff c \symderiv{1} c_t\\
		\tau(c) = (p,{}\cdot{},\tau(c_f)) &\iff c \symderiv{0} c_f
	\end{align*}
	where we abbreviate $c=\config{ N, \aff, \pc}$, $c_t=\config{N', \aff', p \land \pc}$, and $c_f=\config{N', \aff', \lnot p \land \pc}$
\end{sketch}
We call the structures resulting from $\tau$-transformation \emph{Typed Affine Decision Trees} (TADT).
A TADT inherits the type from its underlying algebra of typed affine functions $\affset(n,m)$ (cf., \cref{theo:affine_operations,prop:typed-algprop}).
Similar to ADTs, TADT can also be generalized to acyclic graph structures:
\begin{definition}[Typed Affine Decision Structure]
	\label{def:tads}%
	An ADS over the algebra  $(\affset(n,m), +_t, \cdot, \circ_t) $ where all predicates 
	are linear inequalities in $\realset^n$
	is called \emph{Typed Affine Decision Structure} of type $n \times m$. 
  
	The set of all such decision structures is denoted by $\tadsset(n,m)$, and the set of all typed affine decision structures of any type with:
	\[
	\tadsset = \bigcup_{ n, m \in \natset^+ }\tadsset(n, m)
	\]
\end{definition}
TADS are special kinds of ADS\@. Thus, they inherit the ADS semantics (cf.,\ \cref{def:ads_sem}) when specializing $\Sigma$~to~$\realset^n$ and $\sigma$~to~$\vec x$.
The fact that the semantics of leafs is given by affine functions that are also applied to $\vec x$ is not important for the resulting specialized definition which reads:
\begin{definition}[Semantics of TADS]
	\label{def:tads_sem}%
	The semantic function 
	\[ \semtads{}\colon \tadsset(\inn,\out) \rightarrow (\realset^{\inn} \rightarrow \realset^m) \]
	for TADS is inductively defined as
	\begin{align*}
	 	\semtads{\alpha}(\vec x) &\defined \alpha(\vec x) \\
	 	\semtads{(p, l, r)}(\vec x) &\defined
	\begin{cases}
	  \semtads{l}(\vec x) &\textrm{if } \llbracket p \rrbracket(\vec x) = 1 \\
	  \semtads{r}(\vec x) &\textrm{if } \llbracket p \rrbracket(\vec x) = 0 
	\end{cases}
	\end{align*}
\end{definition}
Every PLNN $N$ defines an ADT $t_N$ over $\affset$. We can therefore apply the results of \cref{sec:ADS}. In particular, 
we can apply the various reduction techniques, which transform $t_N$ into the more general form of an ADS, or more precisely, of
a TADS\@. 

Optimizations in terms of semantic reduction and infeasible path elimination do not alter the semantics of
a (T)ADS. In other words 
\[ \tadsset(N) \ = \ \set[\big]{ t \given  \semtads {t} =  \semtads{(\tau(N)} } \]
is closed under semantic reduction and infeasible path elimination.
Moreover, we have:
\begin{theorem}[Correctness of $ \semtads {t}$]
	\label{def:TADT-correctness}%
	Let $N \in \plnnset$ and $t \in \tadsset(N)$.
	Then we have:
	\[ \semds{N} = \semtads {t} \]
\end{theorem}
In the following, we sometimes abuse notation and also write $\tau(N)$ for other members of 
$t \in \tadsset(N)$ when the concrete structure of the TADS does not matter. This concerns, in particular, \cref{sec:showcase}
where we always apply semantic reduction and infeasible path elimination to reduce size.

\noindent
Following~\cite{Guidotti2018}:
\begin{displayquote}[\cite{Guidotti2018}]
	In the state of the art a small set of existing interpretable models is recognized: decision tree, rules,
	linear models [$\ldots$]. These models are considered easily understandable and interpretable
	for humans.
\end{displayquote}
we have:
\begin{corollary}[Model Explanation]
	\label{theo:ModelExpl}%
	TADS provide precise solutions to the \emph{model explanation} problem, and therefore also to the  \emph{outcome explanation} problem.
\end{corollary}
Please note that outcome explanation is easily derived from model explanation simply by following 
the respective evaluation path.
\begin{figure}
\centering
\includegraphics[scale=.5]{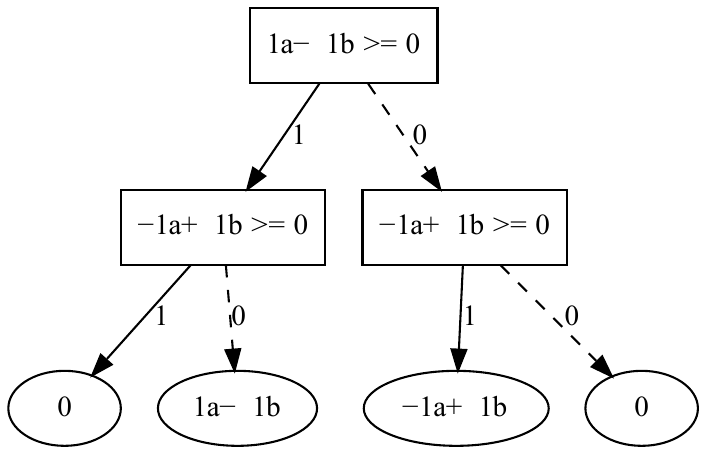}
\caption{The TADS $\nettotads({N_{*}})$.}
\label{XORTADS}
\end{figure}

\begin{example}[XOR-TADS]
	As an example, the resulting TADS of the symbolic execution ADS of $N_{*}$ is shown in \cref{XORTADS}.
\end{example}

\subsection{The TADS Linear Algebra}
\label{sec:tads-algebra}%
\begin{figure*}[!ht]
	\centering
	\subfloat[(T)ADS $2\liftedsmul\relu{2}{1}$]{
		\includegraphics[width=\linewidth,height=80pt,keepaspectratio]{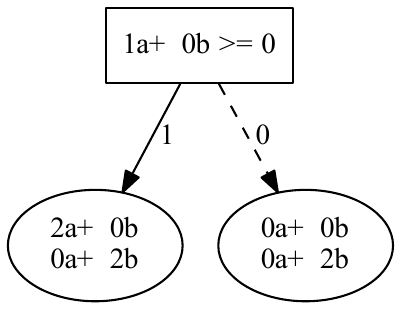}
	} \hfil
	\subfloat[(T)ADS $3\liftedsmul\relu{2}{2}$]{
		\includegraphics[width=\linewidth,height=80pt,keepaspectratio]{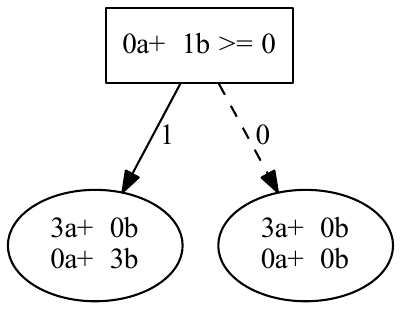}
	} \hfil
	\subfloat[Sum of (a) and (b) $2\liftedsmul\relu{2}{1} \liftedplus 3\liftedsmul\relu{2}{2}$]{
		\includegraphics[width=\linewidth,height=100pt,keepaspectratio]{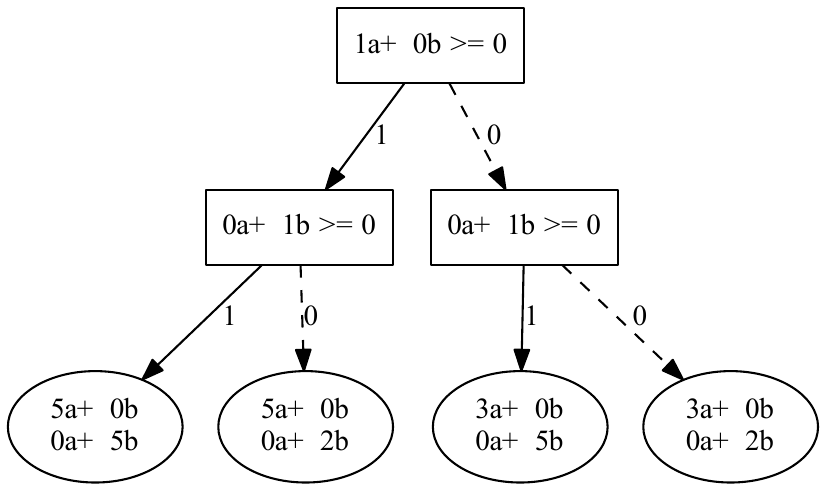}
	}
	\caption{Example for (T)ADS addition. The (T)ADS in (a) and (b) are based on partial ReLUs and (c) is the sum of both. The input vector is given as $\vec x = (a,b)$ with $a,b\in\realset$.}
	\label{fig:addition}
\end{figure*}
According to \cref{prop:typed-algprop}, $(\affset, +_t, {}\cdot_t{})$ forms a typed algebra.
Moreover, due to the canonical representation of affine functions, $\affset$ also supports the
equality relation $=$. Applying \cref{theo:lifted_operators_correctness}, all these operations
can be lifted to obtain the following corresponding operations on TADS: 
\begin{enumerate}[itemsep=2pt]
	\item $\displaystyle \liftedplus \colon \tadsset(n,m) \times \tadsset(n,m) \to \tadsset(n,m) $
	\item $\displaystyle \liftedminus \colon \tadsset(n,m) \times \tadsset(n,m) \to \tadsset(n,m) $
	\item $\displaystyle \liftedsmul \colon \realset \times \tadsset(n,m) \to \tadsset(n,m)$
	\item $\displaystyle \liftedeq \colon \tadsset(n,m) \times \tadsset(n,m) \to \tadsset(n,1) $
\end{enumerate}
%
%Where $t_1 \liftedminus t_2 \defined t_1 \liftedplus \big( (-1) \liftedsmul t_2\big)$.
These operations lift, in the order that they are given, (1)~addition, (2)~subtraction\footnote{Subtraction is usually not stated explicitly as it can be defined using addition and scalar multiplication.}, (3)~scalar multiplication, and (4)~equality.
The resulting TADS has size $\bigO(\abs{t_1} \cdot \abs{t_2})$ where $\abs{t_i}$ is the number of nodes of considered TADS $t_i$.
An example of addition is given in \cref{fig:addition}.

The operations $\liftedplus$ and $\liftedsmul$ are characteristic for vector spaces. 
Indeed, TADS form a (function) vector space (cf.,\ \cref{prop:algprop,prop:typed-algprop}):
\begin{theorem}[The TADS Linear Algebra]
	The triple $(\tadsset, \liftedplus_t, \liftedsmul_t)$ forms a typed linear algebra.
\end{theorem}
We will exploit this theorem in \cref{sec:showcase}.

However, when lifting these two operators over affine predicates, a second interpretation occurs naturally: that of \emph{piece-wise affine functions}.
Both interpretations are compatible, as stated in the following lemma.
\begin{theorem}[Two Consistent Views on TADS]
	Let $\paff_1,\paff_2 \colon \realset^n\to\realset^m$ be two piece-wise affine functions and $\vec x \in \realset^n$ be a real vector.
	Define $\aff_1$ as the affine function of $\paff_1$ that is associated with the region for $\vec x$ and $\aff_2$ for $\paff_2$, respectively and denote with $\genericoperator$ a generic operation over (piece-wise) affine functions.
	Then, if for all such $\vec x$
	\[ \paff_1(\vec x) \genericoperator \paff_2(\vec x) = \aff_1(\vec x) \genericoperator \aff_2(\vec x) \]
	holds both interpretations agree for $\genericoperator$.
\end{theorem}
One can easily show that this is indeed the case for $\genericoperator\in\set{\liftedplus,\liftedminus,\liftedsmul}$.
However, there is a slight difference in the interpretations.
The first \emph{lifts} affine functions over affine predicates and the latter \emph{associates} affine functions with affine predicates.
This distinction can, for example, be seen in the signature of the respective semantics:
\begin{alignat*}{3}
	&\adssem{t} &&\colon \realset^n \to (&&\realset^n \to \realset^m) \\
	&\tadssem{t} &&\colon \realset^n \to &&\realset^m
\end{alignat*}
For TADS to be equivalent to piece-wise affine functions, the semantics have to be adapted to $\tadssem{}$, which slightly differs from $\adssem{}$ in that the leafs are also evaluated under the input.

Considering \cref{theo:path_condition_partition}, one can easily see that every path in a TADS defines a polyhedron and that the set of all paths partitions $\realset^n$.
As all terminals of TADS are affine functions, it is straightforward to prove that for every TADS $t$ the semantics $\tadssem{t}$ is a piece-wise affine function.

The complexity of piece-wise affine functions is commonly defined as the smallest number of classes (so-called regions) that are needed to partition the input space \cite{montufar2014number,pascanu2013number,hanin2019complexity}, and which we call \emph{region complexity}. This complexity measure can easily be adopted for TADS using the above reasoning, as it is simply the number of all paths from the root to its terminals. In other words, TADS are linear-size representations of PAF with respect to their region complexity, which implies:
\begin{theorem}[Complexity of Operations]
	\label{theo:ComplO}%
	The operations $\liftedplus, \liftedminus, \liftedeq$ are of quadratic and $\liftedsmul$ of linear time region complexity.
\end{theorem}
\begin{sketch}
	Via structural induction along \cref{def:lifted_operators} it is easy to establish that each node of the tree underlying $t_1$ 
	is processed at most once, while the nodes of the tree underlying $t_2$ 
	may be processed at most once for each leaf of $t_1$. The theorem follows from the fact that the number of nodes in a binary tree is at most twice the number of its paths.
\end{sketch}
Interesting is the expression $t_1 \liftedminus t_2$ which evaluates to the constant function $0$ iff  $t_1$ and $t_2$ are semantically equivalent. 
Thus, we have the following: 
\begin{corollary}[Complexity of $\equiv$]
	Deciding semantic equivalence between two TADS has quadratic time region complexity.
\end{corollary}
Please note that this way of deciding semantic equivalence does not only provide a 
Yes/No answer, but, in case of failure, also precise diagnostic information:
For $t_2 - t_1$ we have (see \cref{minus})
\begin{itemize}
	\item positive parts mark regions where $t_2$ is bigger
	\item zero marks regions where both TADS agree
	\item negative parts mark regions where $t_1$ is bigger
\end{itemize}
This is particularly interesting when combined with a threshold $\varepsilon$ (see \cref{fig:xor_difference_function_epsilon}).

\subsection{The TADS Typed Monoid}
\label{sec:tads_monoid}

As shown in previous sections, TADS are a comprehensible and efficient representation of piece-wise affine functions. In the following, we will go even further and show that TADS also directly support all common operations on piece-wise affine functions.

Piece-wise affine functions form a typed monoid under function composition, i.e., the composition of two piece-wise affine functions is again piece-wise affine, assuming that domain and co-domain adequately match. This property is highly useful both for the design of neural networks (which are themselves fundamentally compositions of multiple, simple piece-wise affine functions) and neural network analysis, as will be shown in \cref{sec:showcase}.

Consider the following result, which follows as a consequence of the previous correctness theorems and the compositionality of 
$\semds{\cdot}$:
\begin{corollary}[Compositionality]
	\label{theo:compositionality}%
	Let $N_0, N_1, N_2 \in \plnnset$ with $N_0 = N_1\netcomp N_2$ and $t_i \in \tadsset(N_i)$.  Then we have:
	\begin{align*}
		\semds{N_0} &= \semds{N_1\netcomp N_2} \\
		&= \semds{N_2} \circ \semds{N_1} \\
		&= \semtads{t_2} \circ \semtads{t_1} =
		\semtads{t_0}
	\end{align*}
\end{corollary}
Obviously, there is a gap in the result that poses the question: 
\enquote{Is it possible to define composition operator that directly works on TADS?}
Just composing
the affine functions at the leafs, which would be sufficient to, e.g., for $\liftedplus$, is insufficient because of the side effect
of the first TADS\@.
Thus, we end up with the following composition operator that
handles this side effect in a way that is typical for structured operational semantics:
\begin{definition}[TADS Composition]
	\label{def:affinis_join}%
	The composition operator $\tadsjoin$ of TADS with type
	\[ \tadsjoin\colon\tadsset(n,r) \times \tadsset(r, m)\to\tadsset(n,m) \]
	is inductively defined as
	\begin{align*}
		\aff \tadsjoin \aff' &= \aff' \circ \aff \\
		\aff \tadsjoin (p,l,r) &= (p \circ \aff, \aff \tadsjoin l, \aff \tadsjoin r) \\
		(p,l,r) \tadsjoin t &= (p , l \tadsjoin t, r \tadsjoin t) 
	\end{align*}
	where $\aff,\aff'\in \affset$ are TADS identified with their affine function, $t,l,r\in \tadsset$ are TADS, and $p\in\mathcal{P}$ is a predicate.
	Here $p \circ \aff$ with $p=\aff'(x)_i \geq 0$ is defined as 
	\[ (\aff' \circ \aff)(x)_i \geq 0 \]
\end{definition}
Notice that this definition is similar to the lifted operators of \cref{def:lifted_operators}.
However, TADS composition is not side-effect free as can be seen by the modification of the predicate in the second case.
This is due to the fact that the first TADS distorts the input vector space of the second TADS\@.
Again, let us formalize the correctness of this operation.
\begin{figure*}[!ht]
	\centering
	\subfloat[TADS $2\liftedsmul\relu{2}{1}$]{
		\includegraphics[width=\linewidth,height=80pt,keepaspectratio]{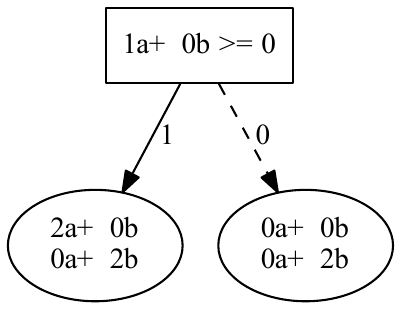}
	} \hfil
	\subfloat[TADS $3\liftedsmul\relu{2}{2}$]{
		\includegraphics[width=\linewidth,height=80pt,keepaspectratio]{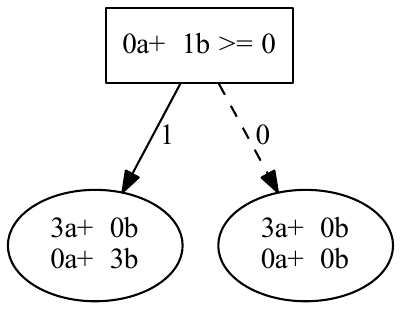}
	} \hfil
	\subfloat[Composition of (a) and (b) $2\liftedsmul\relu{2}{1} \tadsjoin 3\liftedsmul\relu{2}{2}$]{
		\includegraphics[width=\linewidth,height=100pt,keepaspectratio]{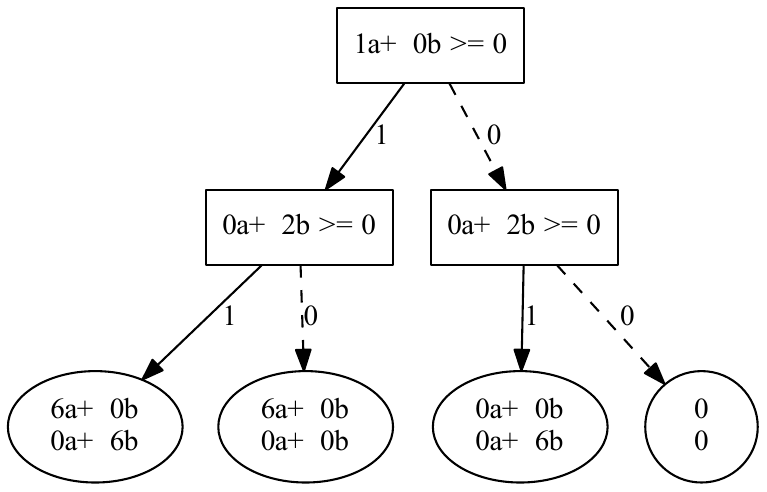}
	}
	\caption{Example for TADS composition. The TADS (a) and (b) are based on partial ReLUs. TADS (c) is the composition of (a) and (b). Note the difference between \emph{lifting} (\cref{fig:addition}) and \emph{composing} in the inner nodes. The input vector is given as $\vec x = (a,b)$ with $a,b\in\realset$.}
	\label{fig:composition}
\end{figure*}
\begin{theorem}[TADS Composition]
	\label{theo:composition}
	Let $t_1 \in \tadsset(n,r)$ and $t_2 \in \tadsset(r,m)$. Then we have:
	\[ \semtads{t_1 \tadsjoin t_2} = \semtads{t_2} \circ \semtads{t_1} \]
\end{theorem}
\begin{sketch}
	Structural induction along the second
	component and in the inductive step induction along the first component.
\end{sketch}
An example of a composition can be found in \cref{fig:composition}.
\begin{figure*}[!ht]
	\centering
	\subfloat[identity $\varepsilon$]{
		\includegraphics[width=\linewidth,height=40pt,keepaspectratio]{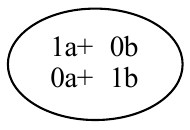}
	} \hfil
	%\subfloat[affine function]{
	%	\includegraphics[width=\linewidth,height=40pt,keepaspectratio]{figures/affine_tads.pdf}
	%} \hfil
	\subfloat[affine function with bias]{
		\includegraphics[width=\linewidth,height=40pt,keepaspectratio]{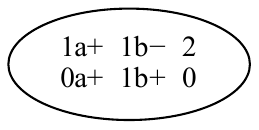}
	} \hfil
	\subfloat[ReLU $\relu{2}{1}$]{
		\includegraphics[width=\linewidth,height=80pt,keepaspectratio]{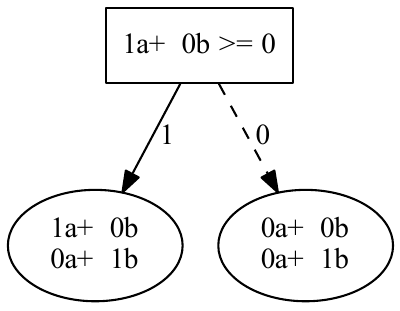}
	}
	\caption{A few examples for atomic TADS.
	The input vector is given as $\vec x = (a,b)$ with $a,b\in\realset$.}
	\label{fig:builing_blocks}
\end{figure*}
This directly yields:
\begin{corollary}[The TADS typed Monoid]
	\label{theo:tads-monoid}%
	The pair $(\tadsset, \tadsjoin) $ forms a typed monoid,
	i.e., an algebraic structure that is closed under type-correct composition and that has
	typed neutral elements $\varepsilon$.
	
	On this structure $\tau$ is a homomorphism between the monoids $(\tadsset, \tadsjoin)$ and $(N, ;)$, i.e., the following diagram commutes
	\begin{center}
		\begin{tikzcd}
			N^2 \ar[r, ";"] \ar[d,"\tau"] & N \ar[d, "\tau"] \\
			\tadsset^2 \ar[r, "\Join"] & \tadsset
		\end{tikzcd}
	\end{center}
\end{corollary}
Due to their similarity to the lifted operators, it is easy to show that composing to TADS results in a third TADS that has size complexity equal to product of its inputs and whose complexity with respect to the measure of affine regions is quadratic in its inputs. 
Following the same line of reasoning as for \cref{theo:ComplO} yields:
\begin{theorem}[Complexity of Composition]
	\label{theo:ComplOp}%
	TADS compositions $\tadsjoin$ has quadratic time region complexity.
\end{theorem}
One may argue that semantic equivalence between two TADS is of limited practical value,
in particular, as in most applications of neural networks, small errors are, to a certain degree accepted. 
In contrast, $\epsilon$-similarity, i.e., whether two TADS differ more than $\epsilon$ 
for some small threshold $\epsilon\in\realset$, can be regarded as a practically very relevant
notion, in particular, to study robustness properties. The corresponding property required for TADS leaves 
is easily defined:
\begin{align*}
	l_\epsilon(x,y) &\defined (\abs{x-y} -\epsilon) \, \indicatorfunc(\abs{x-y}\geq \epsilon) 
\end{align*}
Intuitively, this function yields $0$ if the difference of $x$ and $y$ is less than $\epsilon$ and the absolute value (minus epsilon) of their difference otherwise.
$l_\epsilon$ can easily be realized using only standard algebraic operations and ReLU applications, which are already defined:
\[ l_\epsilon = \Relu( x-y-\epsilon)+\Relu( y-x-\epsilon)\]
Just lifting this function to the TADS level 
\[ t_4 = \Relu( t_1 \ominus t_2 \ominus \epsilon) \oplus \Relu( t_2  \ominus  t_1  \ominus \epsilon)  \]
(where $\Relu(t) = t \tadsjoin \nettotads(\Relu)$)
is sufficient to decide $\epsilon$-similarity. Thus, we have:
\begin{corollary}[Deciding $\epsilon$-similarity]
	\label{cor:Epsilon}%
	$\epsilon$-similarity has quadratic time region complexity.
\end{corollary}
Please note that, again, this way of deciding $\epsilon$-similarity does not only provide a 
Yes/No answer, but, in case of failure, also precise diagnostic information.
All this will be showcased in \cref{sec:showcase}.

In the remainder of this section, we elaborate on the compositionality that is imposed by 
$\tadsjoin$:

\begin{corollary}[Layer-wise Transformation]
	By \cref{theo:composition}, we can transform a PLNN layer-wise into a TADS.
	\begin{align*}
		\semds{N} &= \semds{\aff_1 \netcomp \ldots \netcomp \aff_n} \\
		&=\semtads{\nettransform(\aff_n)} \circ \dots \circ \semtads{\nettransform(\aff_1)} \\
		&= \semtads{\nettransform(\aff_1) \tadsjoin \cdots \tadsjoin \nettransform(\aff_n)} \\
		&= \semtads{\nettransform(N)}
	\end{align*}
\end{corollary}
As a consequence, the transformation function $\tau$ can also be inductively defined
using the following three atomic TADS
\begin{itemize}
	\item identity $\varepsilon$
	\item affine functions $\aff\colon\realset^n\to\realset^m$ where $n,m\in\natset$
	\item single ReLUs $\relu{n}{i}$ where $n,i\in\natset$, $i \leq n$
\end{itemize}
which are illustrated in \cref{fig:builing_blocks}.

\begin{corollary}[Inductive Definition of $\tau$]
	The transformation of a network to a TADS can be defined inductively as
	\begin{align*}
		\tau'(\varepsilon) &= \varepsilon \\
		\tau'(\alpha \netcomp N) &= \alpha \tadsjoin \tau'(N) \\
		\tau'(\relu{k}{i} \netcomp N) &= \big(\,\dotprod{\unitvec{i}}{\vec x} \geq 0, \idmat{k}, \idzmat{k}{i}\,\big) \tadsjoin \tau'(N) \\
		\shortintertext{such that}
		\tau'(N) &= \tau(N) .
	\end{align*}
\end{corollary}
This consistency of viewpoints and operational handling indicates that the TADS setup is natural, and that it supports to approach PLNN analysis and explanation from various perspectives 

\section{TADS at Work}
\label{sec:showcase}%
\begin{figure*}[htbp]
	\centering
	\subfloat[Graph of $N_1$]{
		\includegraphics[width=\linewidth,height=150pt,keepaspectratio]{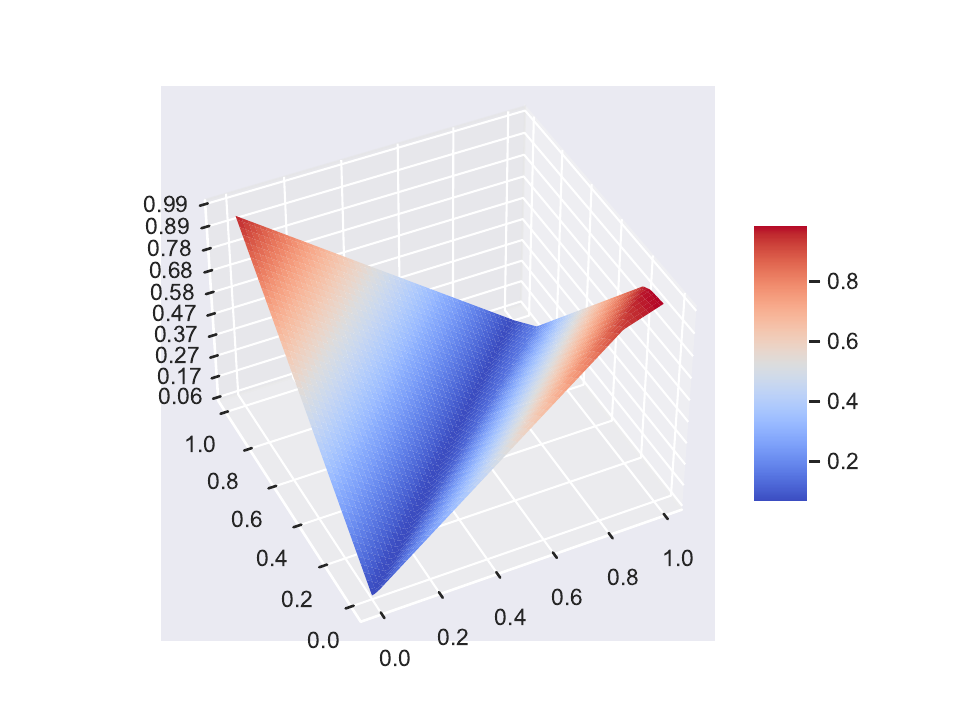}
		\label{fig:p1func}
	} \hfill
	\subfloat[Graph of $N_2$]{
		\includegraphics[width=\linewidth,height=150pt,keepaspectratio]{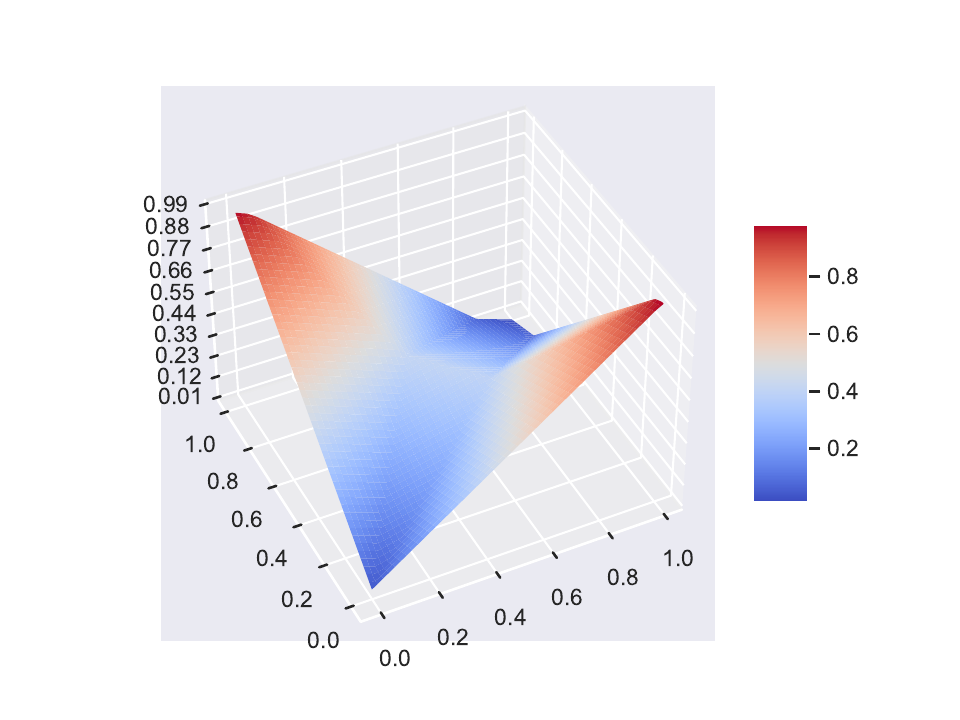}
		\label{p2func}
	}
	\caption{Function graphs corresponding to the PLNNs $N_1$ and $N_2$. Observe that both PLNNs fulfill the conditions of the XOR problem very closely.}
\end{figure*}

\begin{figure*}[htbp]
	\centering
	\subfloat[TADS of $N_1$]{
		\includegraphics[width=.9\linewidth]{"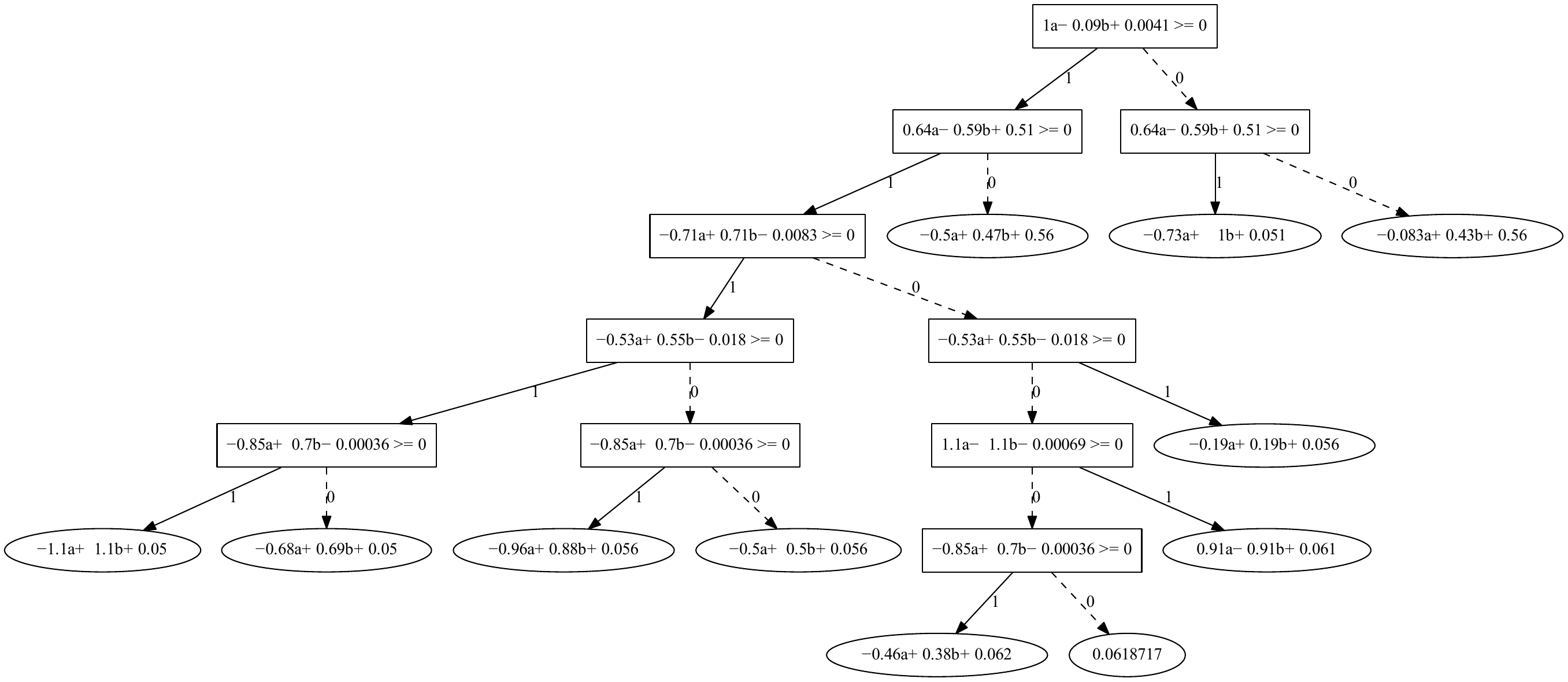"}
		\label{p1tads}
	} \hfill
	\subfloat[TADS of $N_2$]{
		\includegraphics[width=.9\linewidth]{"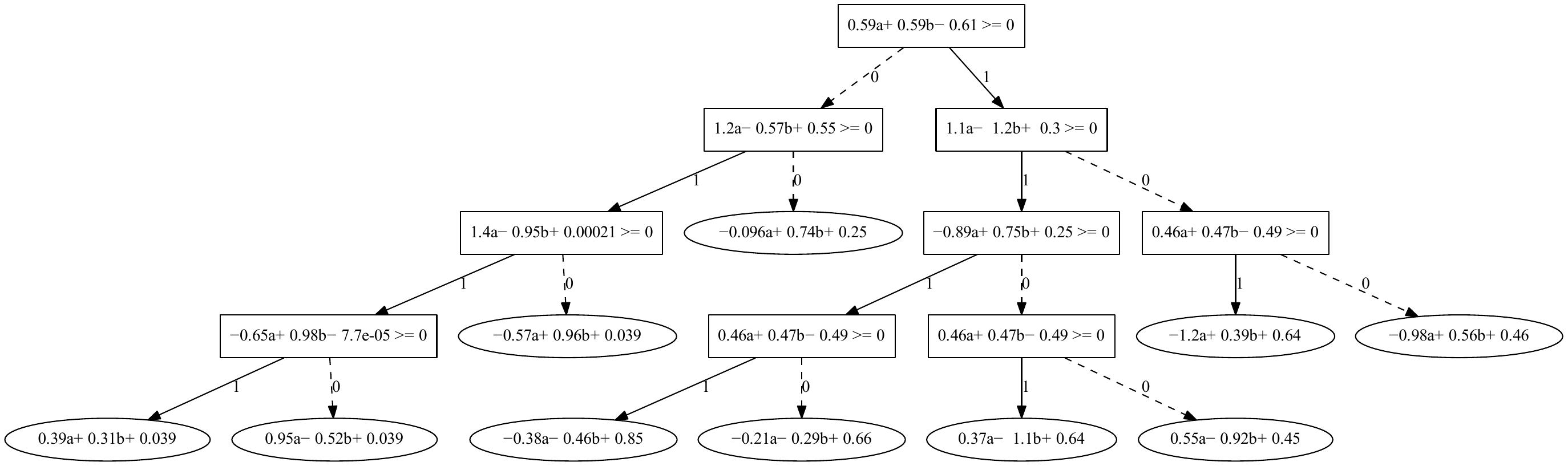"}
		\label{p2tads}
	}
	\caption{TADS corresponding to the PLNNs $N_1$ and $N_2$. Note that both TADS are a full characterization of the semantic functions $f_1$ and $f_2$, respectively.}
\end{figure*}
In this section, we continue the discussion of  the XOR function as a minimal example to showcase the power of TADS for:
\begin{itemize}
\item \textbf{Model Explanation.} For a given PLNN, describe precisely its behavior in a comprehensible manner. This allows for
a semantic comparison of PLNNs comprising (approximative) semantic equivalence
with precise diagnostic information in case of violation.

\item \textbf{Class Characterization.} PLNNs are frequently extended by the so-called argmax function to be used as classifiers. TADS-based
class characterization allows one to precisely characterize the set of inputs that are specifically classified, or the set of inputs
where two (PLNN-based) classifier differ.

\item \textbf{Verification.} Verification is beyond the scope of this section but will be discussed in~\cite{STTT4}
in the setting of digit recognition.
\end{itemize}
In the remainder of this section we focus on the impact of Model Explanation and Class Characterization.
Two properties of TADS are important here:

\textbf{Compositionality.} Due to the compositional nature of TADS, any TADS that represents a given PLNN can be modified and extended by \textit{output interpretation} mechanisms. This mirrors a very important use case of neural networks; while neural networks are fundamentally functions ${\realset^{\inn} \rightarrow \realset^{\out}}$, they are often used for discrete problems, which requires a different interpretation of their output. 

\textbf{Precision.} As the TADS transformation of a PLNN is semantics-preserving, all results are precise. 
 
Based on these properties, it is possible to solve  all the aforementioned problems elegantly by simple algebraic transformations of TADS.

\subsection{Model Explanation and Algebraic Implications}
\begin{figure}[htbp]
\centering
\includegraphics[scale=.3]{"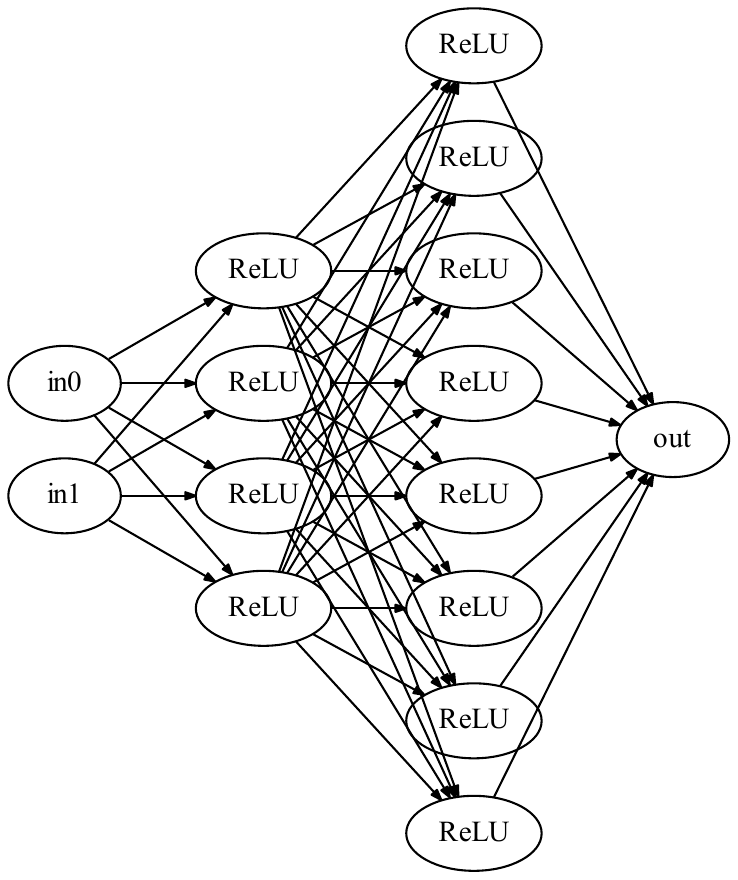"}
\caption{The architecture of the networks $N_1$ and $N_2$. Weights and biases are omitted for brevity.}
\label{fig:n1architecture}
\end{figure}

To start, we train a small neural network to solve the continuous XOR problem. The resulting network, $N_1$, represents a continuous function $f_1 = \netsem{N_1}$ (see \cref{fig:p1func}). $N_1$ solves the XOR problem relatively well, with all corners being within a distance of $<0.1$ to the desired values of $1$ and $0$ respectively.

The architecture of $N_1$ is shown in \cref{fig:n1architecture}. Note that this architecture is much bigger than the architecture for $N_*$ (cf., \cref{sec:PLNN}). This is needed as the training procedure is approximate and does not reach a global optimum. On all substantially smaller networks, we failed to train a network that was close to the specifications of XOR. 

\subsubsection{Model Explanation}
First, we consider full model explanation of $N_1$.\footnote{Of course, in this two-dimensional case, a function plot akin to \cref{fig:p1func} might seem sufficient, but this is not feasible in anything beyond two-dimensional problems.} We can attain a precise and complete characterization of $f_1$ by creating the corresponding TADS $t_1 = \nettotads(N_1)$, as shown in \cref{p1tads}. This TADS describes precisely and completely the behavior of $f_1$ in a white-box manner.

Similarly to the function plot shown in \cref{fig:p1func}, the TADS gives a comprehensible view of $f_1$.In contrast to the mere function plot, the TADS of \cref{p1tads} is a solid basis for further systematic analyses and extends to more than two dimensions. 

Model Explanation as illustrated here is the basic use case of a TADS as a white-box model for PLNNs, however, the true power of TADS becomes apparent when used for high-level analyses using algebraic operations on TADS.

\subsubsection{Algebraic Implications}

\begin{figure}[htbp]
	\centering
	\includegraphics[width=\textheight-6cm,height=\columnwidth,keepaspectratio,angle=90]{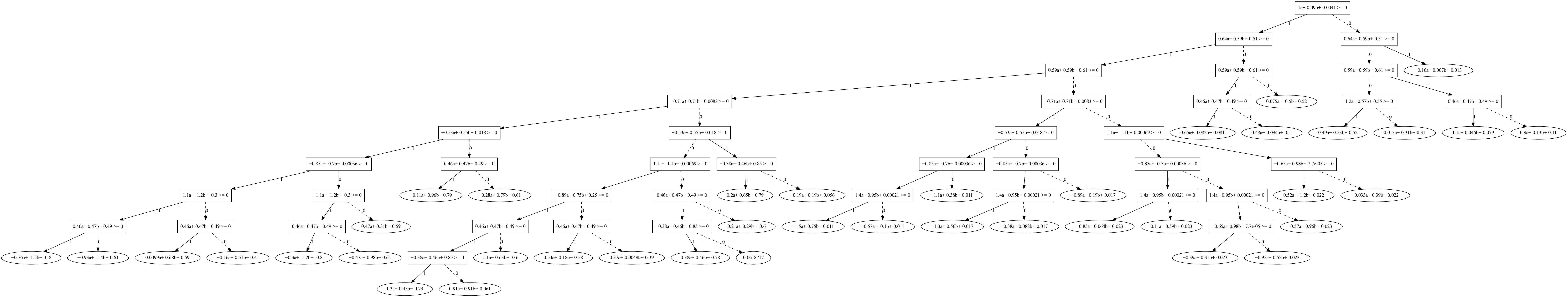}
	\caption{The TADS $t_2-t_1$ describing the difference of $f_2$ and $f_1$. This TADS corresponds to the function plot of \cref{minus}.}
	\label{minustads}
\end{figure}

\begin{figure}[htbp]
	\centering
	\includegraphics[scale=.5]{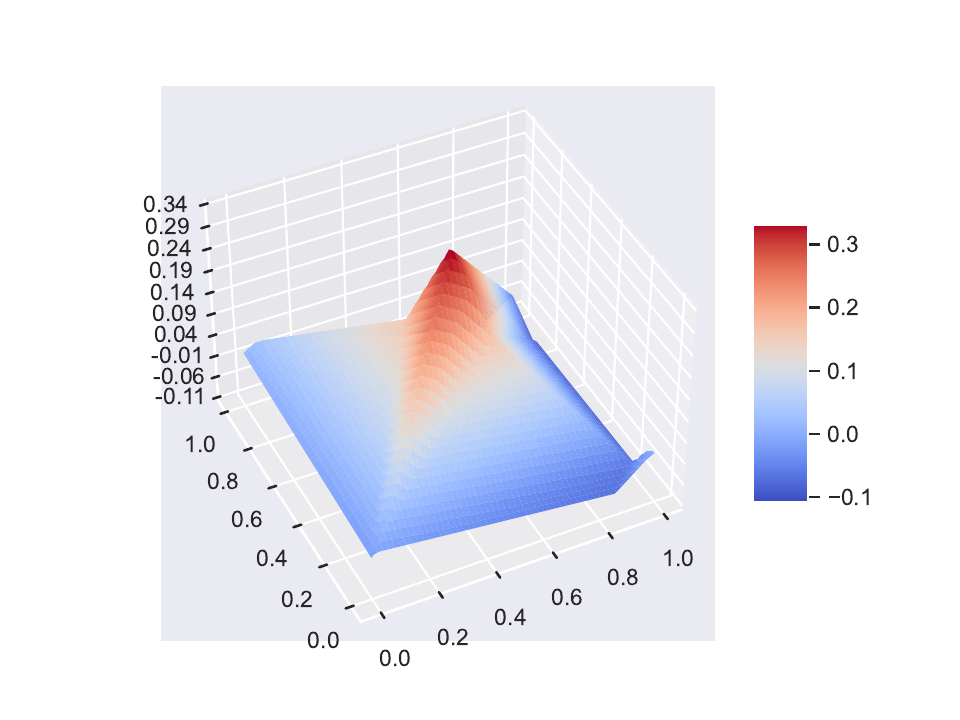}
	\caption{The function plot describing the difference between the two networks $N_1$ and $N_2$.}
	\label{minus}
\end{figure}

As mentioned in the last section, the training process of neural networks is approximate and can lead to many different solutions. A very natural question to ask is: 
\enquote{How differently do two neural networks solve the same problem?}. This question can be answered using algebraic operations on TADS.

Consider $N_2$, a PLNN that has also been trained with the network architecture shown in \cref{fig:n1architecture}, but with a different setting of hyperparameters.\footnote{The discussion of the learning process is beyond the scope of this paper.} 
Its represented (semantic) function $f_2 = \netsem{N_2}$ is depicted in \cref{p2func} and the corresponding TADS $t_2 = \nettotads(N_2)$ in \cref{p2tads}.

As TADS form a linear algebra, one can easily mirror the computation $f_2 - f_1$ by $t_2 - t_1$ on TADS level.
The result is identical because the transformation process is precise, i.e.,
\[ \netsem{N_2} - \netsem{N_1} = f_2 - f_1 = \tadssem{t_2 - t_1} \]
The resulting difference TADS $t_3$ is shown in \cref{minustads} and the corresponding function graph in \cref{minus}. 

The TADS $t_3$ is ideal to study the semantic difference between PLNN $N_1$ and$N_2$. Most interestingly, as can be visually seen in \cref{minus}, the largest differences between both networks occur in the middle of the function domain, i.e., in the region most distant from the edges where the XOR problem is clearly defined. This matches the intuition that points further away from known points are more uncertain under neural network training. 

Further, observe that the difference of both networks yields a TADS of roughly double the size. This moderate increase in size indicates the
similarity of $N_1$ and~$N_2$, as 
linear regions of the difference TADS $t_3$ result from the intersection of the regions for $f_1$ and $f_2$ which could, in the worst case, grow quadratically.

\begin{figure}[htbp]
	\centering
	% -6cm to nudge LaTeX's figure placement to a decent layout
	\includegraphics[width=\textheight-6cm,height=\columnwidth,keepaspectratio,angle=90]{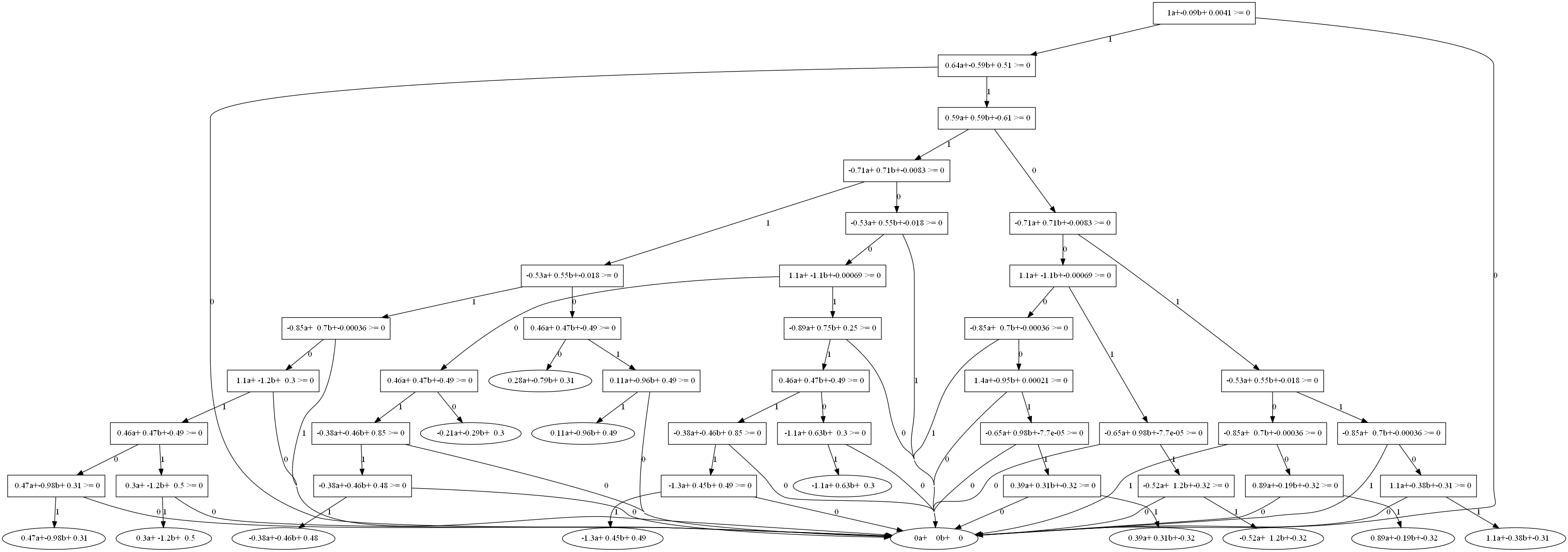}
	\caption{A TADS describing the difference between $f_2$ and $f_1$ iff it exceeds $\epsilon =0.3$.}
	\label{fig:xor_difference_tads_epsilon}
\end{figure}

\begin{figure}[htbp]
\centering
\includegraphics[scale=.5]{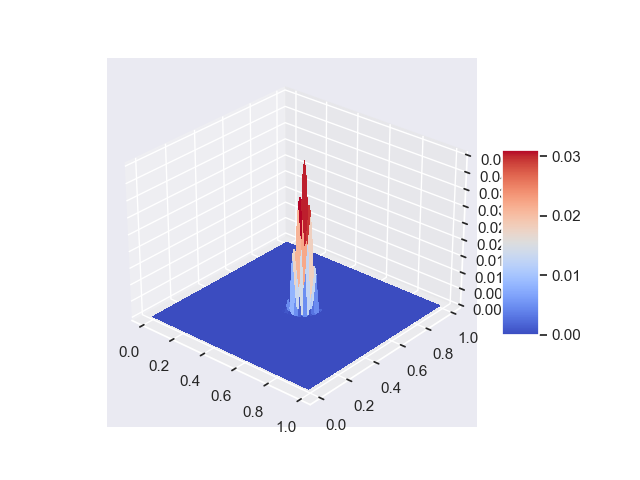}
\caption{The function graph describing the difference between $f_2$ and $f_1$ iff it exceeds $\epsilon =0.3$}
\label{fig:xor_difference_function_epsilon}
\end{figure}

As mentioned above (cf.,\ \cref{cor:Epsilon}), it is also possible to analyse $\epsilon$-similarity via algebraic operations to, in this case, 
obtain the TADS shown in \cref{fig:xor_difference_tads_epsilon}, which is much smaller than the full difference TADS (cf.,\ \cref{minustads}).
The piece-wise affine function of this TADS is visualised in \cref{fig:xor_difference_function_epsilon}.

There are 8 regions in which the difference values exceed $0.3$, all close to the center of $[0,1]^2$. This, again matches the intuition that the volatility of solution grows with the distance to the defined values. 

This result is interesting as it shows that, while the two neural networks that we trained differ, they do not differ more than $0.3$ except for a small region in the center of the input space. 
Similar constructions can be used to analyze robustness of neural networks. Robustness of neural networks is of large interest to neural network research \cite{carlini2017towards} and the application of TADS to this problem is discussed in more detail in~\cite{STTT4}.

\subsection{Classification}

\begin{figure*}[htbp]
	\centering
	\subfloat[TADS $t_1^c$ for the function $g_1 = \indicatorfunc({x \geq 0.5}) \circ f_1$]{
		\includegraphics[width=.9\linewidth]{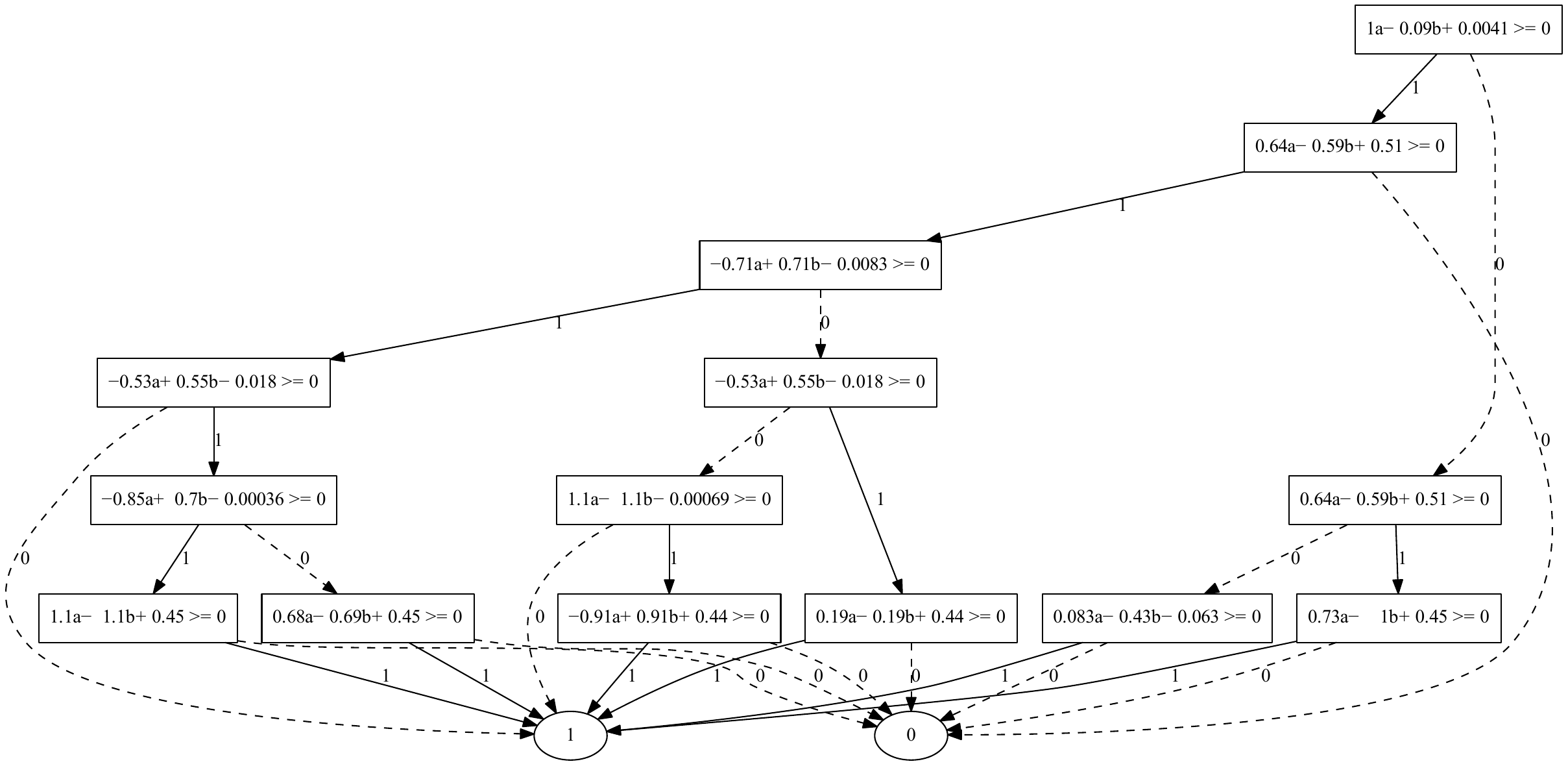}
		\label{output interpreter TADS}
	} \hfill
	\subfloat[TADS $t_2^c$ for the function $g_2 = \indicatorfunc({x \geq 0.5}) \circ f_2$]{
		\includegraphics[width=.9\linewidth]{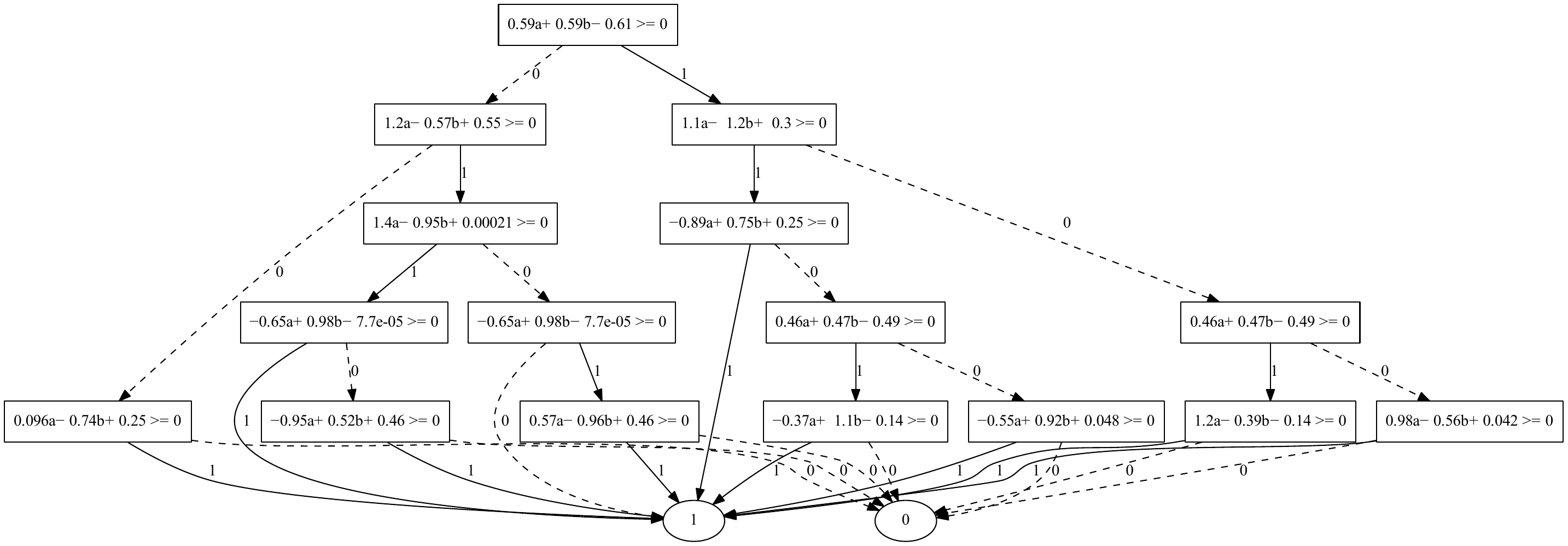}
		\label{output interpreter TADS 2}
	} \hfill
	\caption{Classification TADS that indicates where the PLNNs $N_1$ and $N_2$ output a value greater than 0.5.}
\end{figure*}
\begin{figure}[htbp]
	\centering
	\includegraphics[scale=.5]{"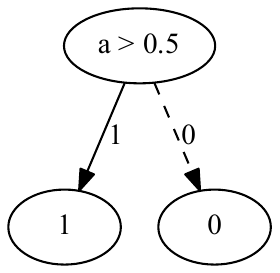"}
	\caption{A TADS describing the function $\indicatorfunc({x \geq 0.5})$ that is used to transform the output of neural networks into discrete values.}
	\label{output interpreter}
\end{figure}

\begin{figure*}[htbp]
	\centering
	\subfloat[${g_1 = \indicatorfunc({x \geq 0.5}) \circ f_1}$ (cf., \cref{output interpreter TADS})]{
		\includegraphics[width=.31\linewidth]{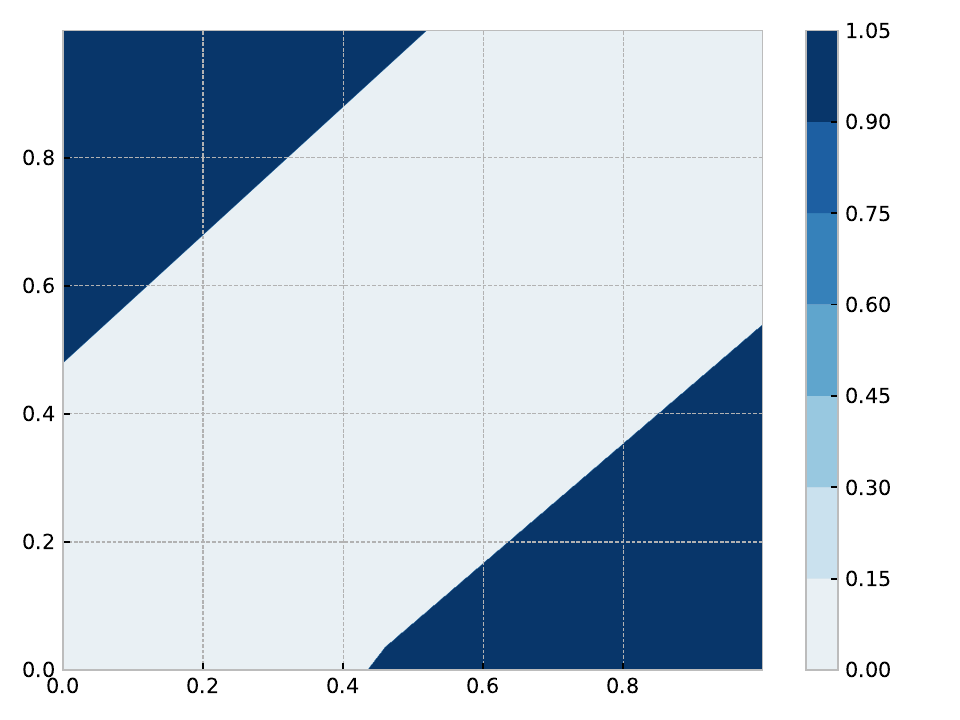}
		\label{bdd plot}
	} \hfill
	\subfloat[${g_2 =  \indicatorfunc({x \geq 0.5}) \circ f_2}$ (cf., \cref{output interpreter TADS 2})]{
		\includegraphics[width=.31\linewidth]{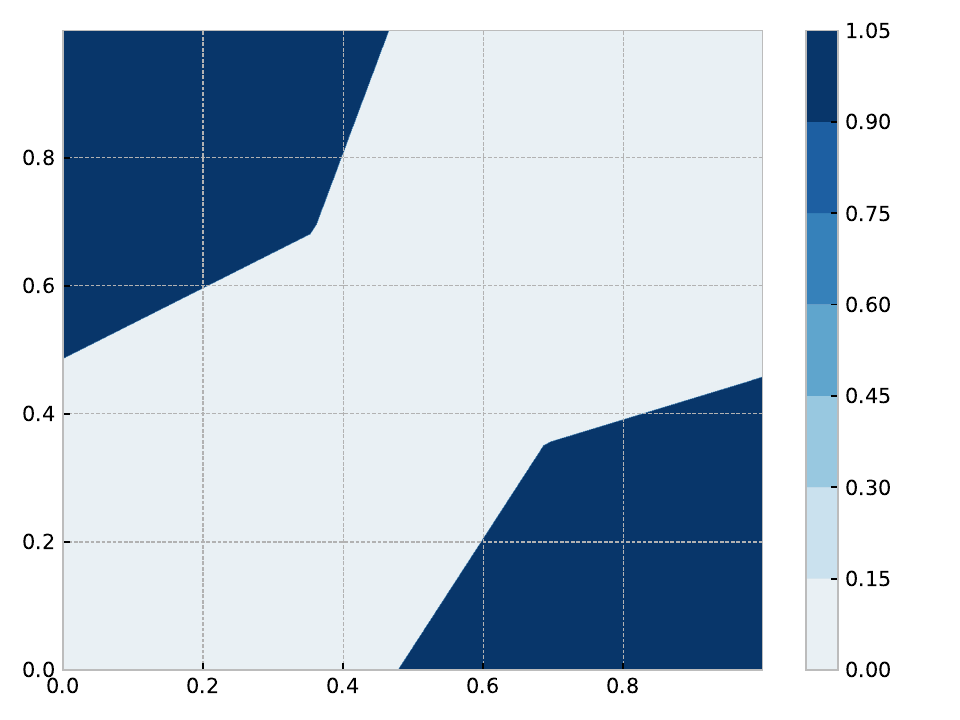}
		\label{bdd plot2}
	} \hfill
	\subfloat[$\tadssem{t_1^c \ \ominus \ t_2^c}$ (cf., \cref{fig:classminustads})]{
		\includegraphics[width=.31\linewidth]{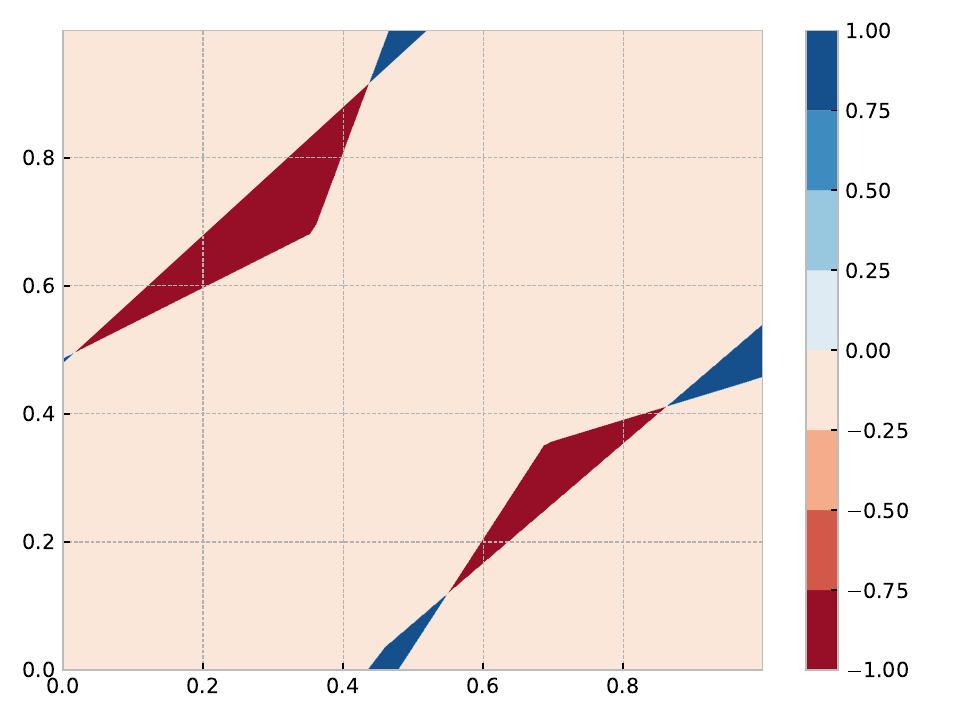}
		\label{classdiffplot}
	} \hfill
	\caption{Contour plots of the classification functions (a) $g_1$ and (b) $g_2$. The difference between both classifiers is visualized in (c). }
\end{figure*}

Applications of neural networks are traditionally split into \emph{regression tasks} and \emph{classification tasks}. In regression tasks, one seeks to approximate a function with continuous values, whereas classification tasks have discrete outputs. As learned, piece-wise linear functions are inherently continuous, classification tasks require an additional step that \emph{interprets} the continuous output of a neural network as one of multiple discrete classes. Note that this is a change of mindset, with the same neural network being interpreted differently depending on the context. 

In our context, one might be interested in a model that classifies each input point $\vec x \in \realset^2$ as either~$1$ or~$0$ instead of returning a real-value.

\subsubsection{Class Characterization}
A standard method for classification tasks is the interpretation of neural network outputs as a probability distribution over classes~\cite{Goodfellow-et-al-2016}. In our XOR example, it is natural to interpret $f_1(\vec x)$ as the probability of~$\vec x$ belonging to class $1$ and $1-f_1(\vec x)$ as the probability of~$\vec x$ belonging to class $0$.

At evaluation time, one might naturally choose the class with the highest probability. Thus, $N_1$'s output is set to~$1$ if it is greater than~$0.5$ and~$0$ otherwise, which is, actually, in line with the definition of~$g_*$. Applying
\[ \indicatorfunc(x \geq 0.5) = 
\begin{cases}
	1 \quad &\text{if } x \geq 0.5 \\
	0 &\text{otherwise}
\end{cases} .\]
to the continuous learned function $f_1$ therefore results in a suitable classifier for this problem:
\[ g = \indicatorfunc(x \geq 0.5) \circ f_1 \]
Note that $\indicatorfunc(x \geq 0.5)$ is not continuous and therefore cannot be represented by a PLNN.\footnote{This is a general observation that holds for all discrete valued classification tasks. Most notably, the argmax function, a standard method for n-ary classification also cannot be represented by a PLNN and must be handled on the side of TADS\@. More discussions on the role of argmax in classification can be found in~\cite{STTT4}.}

To construct the TADS, we use the compositionality of TADS\@.
We manually construct the simple TADS ${\tau(\indicatorfunc(x \geq 0.5))}$, as shown in \cref{output interpreter} and compose it with the TADS $t_1$ of $f_1$.
The resulting TADS 
\[ t_1^c = \tau(\indicatorfunc(x \geq 0.5)) \tadsjoin t_1 \]
is shown in \cref{output interpreter TADS}.
Note that this TADS is reminiscent of a binary decision diagram with just two final nodes. \Cref{bdd plot} and shows precisely which inputs are interpreted as $1$ and which as $0$. As we only have two classes here, this classification can be considered as what is called \emph{class characterization} in~\cite{gossen2021algebraic}
for both classes $0$ and $1$. Please note that class characterizations allows one to change the perspective from a classification task to the task of finding adequate candidates of specific profile, here given as a corresponding class.

This shows that, given an output interpretation that maps the continuous network outputs to discrete classes, it is possible to transform neural networks, fundamentally black-box representations of real valued functions, into semantically equivalent decision diagrams, fundamentally white-box representations of discrete valued functions.

\subsubsection{Comparison of Classifiers}

\begin{figure*}[htbp]
	\centering
	\subfloat[TADS $t_1^c \ \ \liftedeq \ \ t_2^c $]{
		\includegraphics[width=0.48\textwidth]{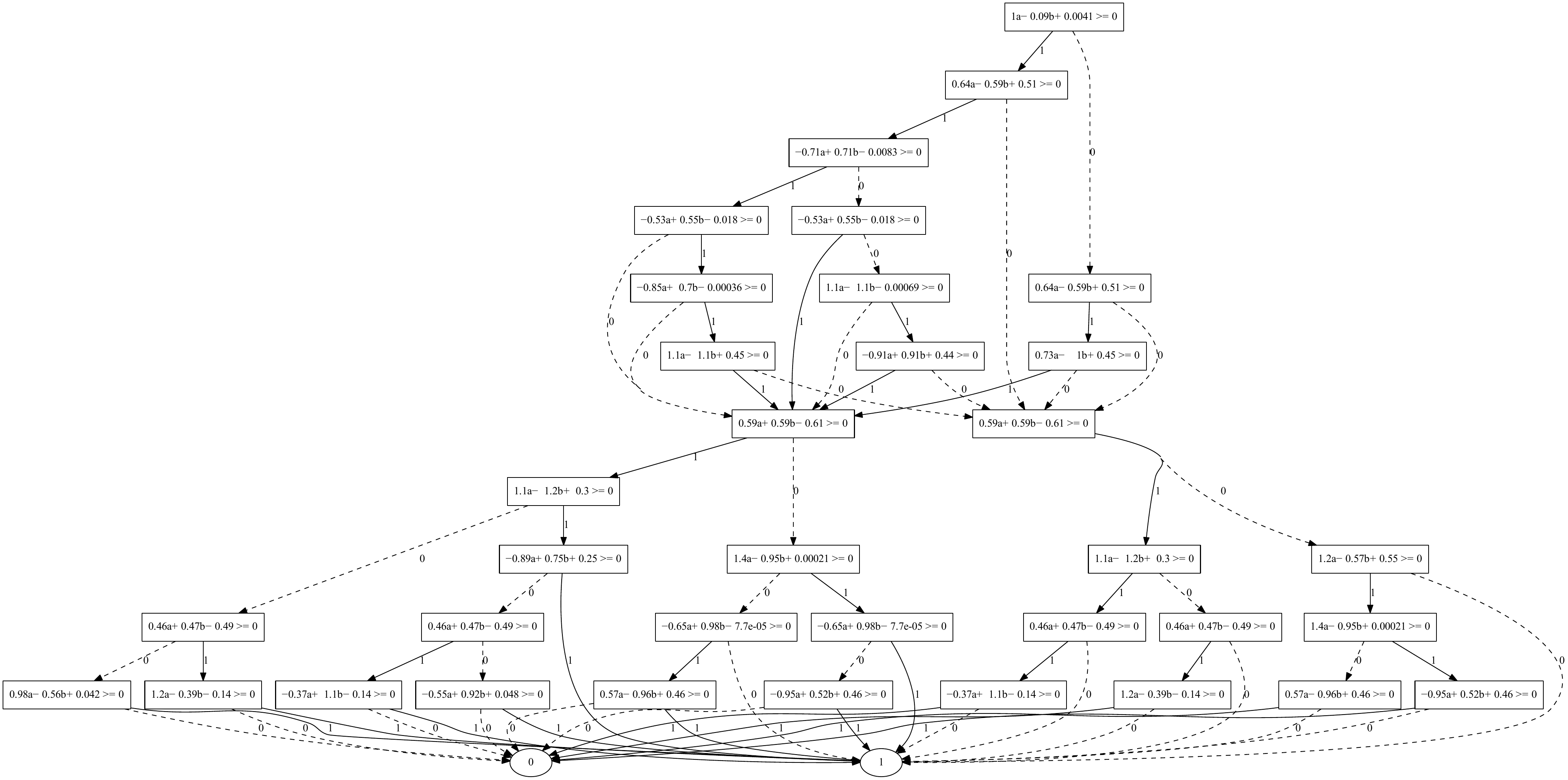}
		\label{classdifftads}
	} \hfill
	\subfloat[TADS $t_1^c \ \ominus \ t_2^c $]{
		\includegraphics[width=0.48\textwidth]{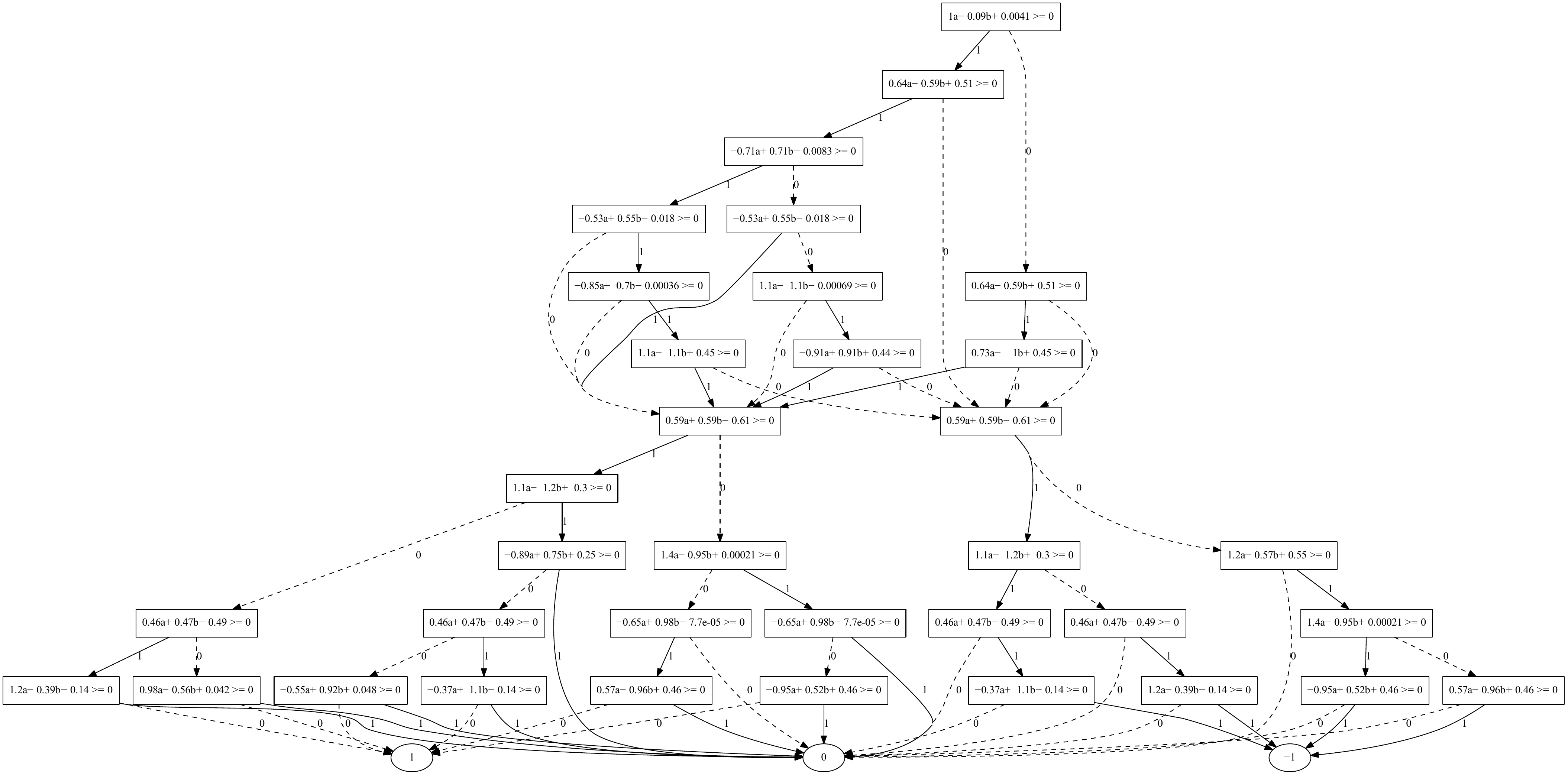}
		\label{fig:classminustads}
	}
	\caption{TADS representing the difference of the classification behavior of $N_1$ and~$N_2$. The TADS in (a) expresses the equality of the classifications while (b) expresses the difference. Thus, by mapping the nodes $-1$ and $1$ to $0$ and $0$ to $1$ one can transform (b) to (a). Note, however, that despite their syntactical similarities, they represent different concepts.}
\end{figure*}

After having constructed TADS that characterize the classification behavior of neural networks, we can also characterize the \emph{difference} in classification behavior of two neural networks.
We can simply do so by using the lifted equality relation to the TADS level and compute the TADS:
\[ t_1^c \ \ \liftedeq \ \ t_2^c \]

The resulting TADS is shown in \cref{classdifftads} and the corresponding function graph in \cref{classdiffplot}. This plot describes precisely the areas where both functions differ and where they coincide. 

Notably, it shows that, while the absolute difference of $f_1$ and~$f_2$ is highest in the center of the interval $[0,1]^2$, the networks agree in that area with respect to classification. Indeed, it appears that the largest difference with respect to classification occurs in the diagonals separating the classes $1$ and~$0$. This is not too surprising, as it is at the borderline between classes were small changes may affect the classification result. 

Using an encoding of boolean values as $1$ and~$0$ respectively, we can also compute the difference of~$t_1^c$ and~$t_2^c$
\[ t_1^c \ \ \ominus \ \ t_2^c \]
This TADS not only describes where $t_1^c$ and~$t_2^c$ disagree, but also how they disagree. The corresponding TADS is shown in \cref{fig:classminustads}.

This shows the utility of TADS for output interpretation. While the absolute difference between two networks is a suitable measure of difference for \emph{regression} tasks, the difference of the classification functions is suitable for \emph{classification}. Playing with this difference, e.g., by modifying the classification function, is a powerful analytical instrument. E.g., in settings with many classes, separately analyzing the class characterizations of the individual classes typically leads to much smaller and easier to comprehend TADS and may therefore be a good means to increase scalability.

\begin{figure}[htbp]
\centering
\vspace*{20pt}
\includegraphics[width=\linewidth]{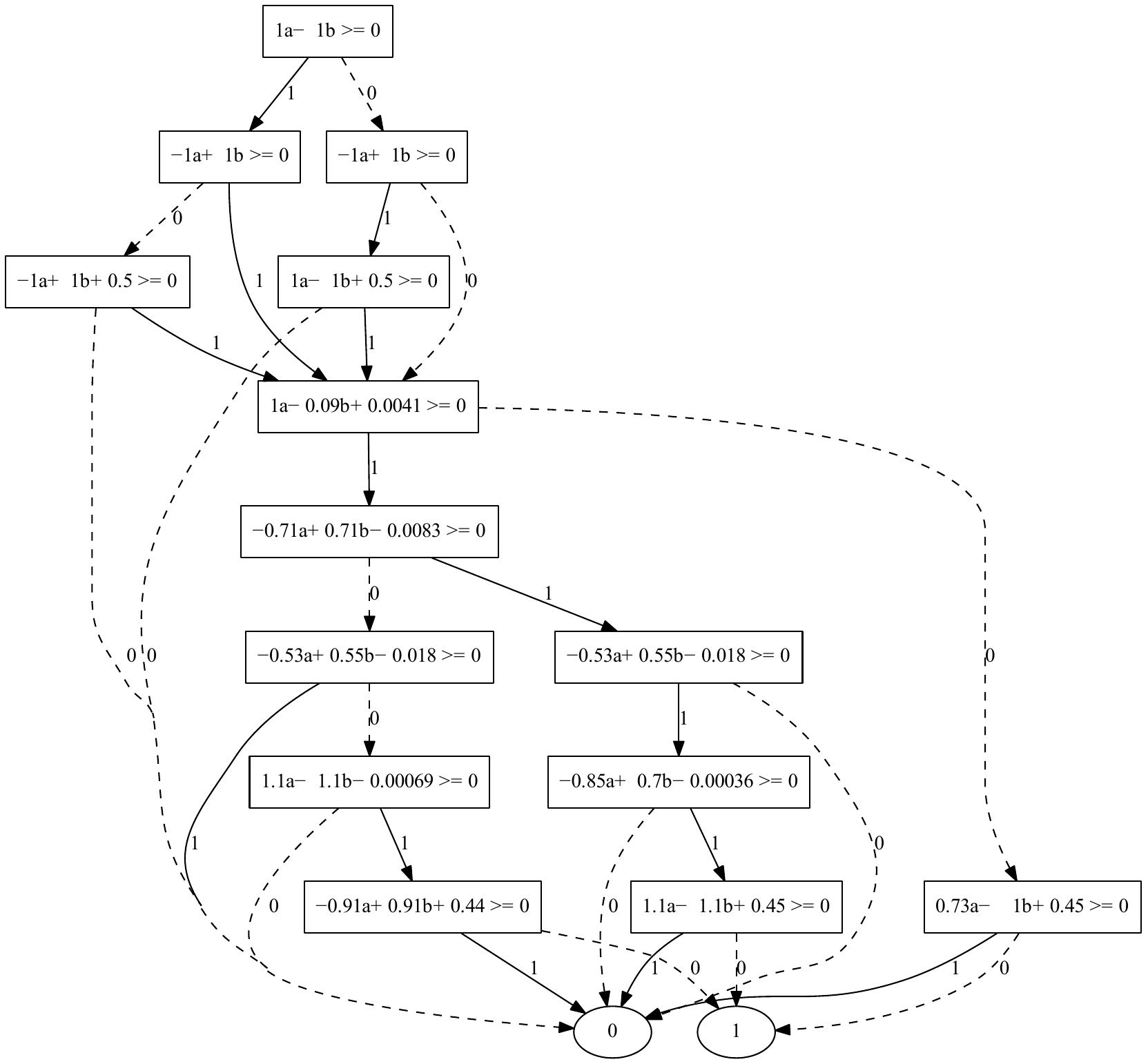}
\caption{The TADS describing the difference between the baseline solution $\indicatorfunc(\abs{x-y} \geq 0.5)$ and the solution learned by $N_1$.}
\label{xor manual tads}
\end{figure}

In machine learning, one often compares learned classifiers to groundtruth solutions by sampling from the groundtruth solution and checking whether the neural network matches the groundtruth predictions. TADS enable a straightforward and precise way of evaluating a neural network in instances where one has access to the groundtruth model. E.g., the TADS of \cref{fig:xor_difference_function_epsilon} precisely specifies where $N_1$ differs from the baseline solution $\indicatorfunc(\abs{x-y} \geq 0.5)$.

\section{Related Work} 
\label{sec:related_work}
The presented TADS-based approach towards understanding of neural networks is explicitly meant to 
bridge between the various existing initiative that aim in the same direction, but typically with quite 
different means. In this section, we review the state of the art under three perspectives: 
\begin{itemize}
\item The intent, explainability, as approached in the neural networks community.
\item Applied concepts, e.g., symbolic execution that aim at (locally) precise results.
\item Applied background, in particular concerning piece-wise affine functions.
\end{itemize}
Whereas the first perspective (\cref{sec:rw/xml}) is conceptually distant, both in its applied technologies as well as 
in its achievements, the mindset of second perspective  (\cref{sec:rw/ac}) is similar in aims and means, but,
except for our previous work, restricts its attention to a locally precise analysis
close to some (partly symbolic) input. The third perspective just concerns the
mathematical background (\cref{sec:rw/mb}).
We are not aware of any previous work that systematically applies algebraic reasoning to achieve 
precise explanation and verification results about neural networks.

\subsection{Machine Learning Explainability}
\label{sec:rw/xml}
In recent years, explainable AI (XAI) as a subfield of machine learning has seen a surge of activity. In line with existing machine learning research, XAI focuses on approaches that scale efficiently at the cost of precision and comprehensibility.

Due to vast amount of work in this direction we can only provide sketch of the field here, which from our perspective is characterized 
by is use of `traditional' deep learning technologies such as gradient based optimization and its focus on directly investigating the neural 
networks themselves in an approximate fashion and without explicit link to some semantic model.

A typical example of a gradient-based method is activation maximization~\cite{mahendran2016visualizing,selvaraju2017grad}, which seeks to find, for one class $C$ and network $N$, the input $\vec x$ for which $\sem{\vec x}$ is maximal for class $C$.
Being based on gradient based optimization, this approach is clearly approximate. 

Other examples of approaches working on the neural network level are frequently found in attribution methods. Attribution methods focus on attributing a prediction $\sem{N}(\vec x) = y$, to parts of the input that one deems responsible for this prediction. In general, this question is unclear and subjective.  As a consequence, there exist multiple different methods that attribute the prediction differently to the original input. Examples include gradient based saliency maps~\cite{simonyan2013deep,mundhenk2019efficient}, layer-wise relevance propagation (LRP)~\cite{bach2015pixel} and deep taylor decomposition~\cite{montavon2017explaining}. As attribution is natural to answer for linear models, these methods focus on linearly approximating the model (gradient based saliency maps) or parts of the model (LRP and deep taylor decomposition). The latter methods depend strongly on the neural network architecture and not on its semantics. 

These methods are useful to gain rough intuition, but they do not offer any guarantees or reliable results. This is a direct consequence of most of these methods working with classical machine learning tools such as backpropagation and numerical optimization, which are fast but approximative~\cite{Goodfellow-et-al-2016}.

The class of XAI methods that is most closely related to TADS are local proxy models like LIME~\cite{ribeiro2016should} and SHAP~\cite{lundberg2017unified}. Both methods consider one fixed input $\vec x$ and treat the model as a black-box. They observe the model's behavior on multiple perturbations of the form $\vec x + p$ with $p$ being sampled randomly. Then, they use simple machine learning models such as a linear classifier or a decision tree to describe the model's behavior on the perturbations $\vec x +p$ they observed. 

These methods are similar in their intent to the TADS approach, as they use conceptually simple models to represent the black-box behavior of a neural network. However, both LIME and SHAP are imprecise. They both sample only a few points $\vec x +p$ in the neighborhood of $\vec x$ and might miss important properties of the neural network model. Further, both methods use a machine learning classifier to represent these points. These classifiers are usually linear models (or a comparably simple model) and cannot capture the full behavior of the network, which leads to potentially large and uncontrolled errors.

We are not aware of XAI methods that provide guarantees strong enough to justify a responsible use in safety critical applications.

\subsection{Conceptually Related Work}
\label{sec:rw/ac}
\paragraph{Symbolic Execution of Neural Networks.} 
More closely related to TADS are approaches to explainability based on symbolic execution of neural networks.

The idea of explainability via symbolic (or rather concolic execution) of neural networks was already explored in the works of~\cite{gopinath2019symbolic}. In their work, the authors translate a given PLNN into an imperative program and concolically execute one given input $\vec x \in \realset^{n}$. This corresponds to exploring the one path of a TADS corresponding to $\vec x$. This yields the path condition and the affine transformation that are responsible for the prediction $\plnn(\vec x)$. The authors further use these explanations to find adversarial examples and find parts of an input that they deem important for a given classification. The results of this work are promising, but (very) local, as they restrict themselves to one linear region of the input.

The authors of~\cite{chu2018exact} propose a method that closely mirrors the method of~\cite{gopinath2018symbolic}.  In essence, both methods are almost identical, but differ in their conceptual derivation of the method. The authors of~\cite{chu2018exact} also consider sets of predictions and work out what features act as discriminators in many of these predictions.

Moving from the idea of explanation, the authors of~\cite{sun2018concolic} consider concolic testing instead. Similar to the work of~\cite{gopinath2018symbolic}, they execute singular inputs concolically. They use the results from concolic execution to heuristically derive new inputs that cover large areas of the input space. 

TADS improve on these approaches in two ways. First, TADS offer a global viewpoint on neural network semantics, independent of a sample set. Second, TADS support algebraic operations on a conceptual level to derive globally precise explanation and verification results. As illustrated in \cref{sec:showcase},
algebraic operations nicely serve as a toolbox to derive tailored and precise analyses.

\paragraph{Neural Network Verification.}
Neural network verification aims to verify properties of neural networks, usually piece-wise linear neural networks using techniques from SMT-solving and abstract interpretation extended by domain-specific techniques~\cite{katz2017reluplex,wang2021beta,tran2019star}. Verification approaches are usually precise, or at least provide a counterexample if a property is shown false. 
Modern solvers can scale quite well, but are still far from being able to tackle practically relevant applications~\cite{bak2021second}. 

Verification approaches are related to TADS as they also provide tools for the precise analysis of piece-wise linear neural networks. However, while SMT-based verification approaches currently scale better than TADS, they focus only on binary answers to a verification problem. They are not able to provide full diagnostics and descriptions of where and how an error occurs. However, please note, SMT-based approaches should not be considered as an alternative but rather as a provider of technologies that can also be applied at the TADS level. In fact, we use SMT solving, e.g., to eliminate infeasible paths in TADS.

\paragraph{Precise Explainability of Random Forests.}
This work is conceptually closely related to and builds upon the work of~\cite{gossen2021algebraic,AlnisSTTT2}. There, ADDs are used and extended to derive explainable global models for random forests. Similar to the approach in this paper, these models are derived through a sequence of semantics-preserving transformations and later on refined by performing algebraic operations on the white-box representation of random forests. In fact, considering the underlying mindset, the work in this paper can be regarded as an extension of our work on random forest to neural networks. However, the much higher complexity of PLNN 
requires substantial generalization, which to our surprise did not clutter the theory, but rather added to its elegance.
\subsection{Technologically Related Work}
\label{sec:rw/mb}
\paragraph{Linear Regions of Neural Networks.}
Vast amounts of research have been conducted regarding the number and shape of linear regions in a given PLNN~\cite{Hinz2021,chu2018exact,montufar2014number,pascanu2013number,zhang2020empirical,Hanin2019,serra2018bounding,hanin2019complexity,Woo2018,Sudjianto2020UnwrappingTB,raghu2017expressive,arora2016understanding}.
Linear regions are of huge interest to neural network research as they give a natural characterization of the expressive power of neural network classes.
This research is beneficial to the understanding of TADS as it can be used to bound the size of TADS and understand where and when explosions and size occur. 
On the other hand, TADS give a precise and minimal representation of the linear regions belonging to given neural network and can be used to facilitate experiments in this field, e.g., to find a linear region containing a negative example for a given property that could not be verified~\cite{katz2017reluplex}.

\paragraph{Structures for Polyhedral Sets.}
At their core, TADS are efficient representations of multiple polyhedral regions within high-dimensional spaces. Similar problems occur in other divisions of computer science, most notably computer graphics.

TADS are closely related to Binary Space Partition Trees (BSP-trees) \cite{Thibault1987} and comparable structures \cite{TondelJB03}. These structures are built to represent a partition of a real-dimensional space into polygons, much like TADS do. TADS extend these structures with optimizations from ADDs to account for domain-specific properties of piece-wise linear functions that are not present in the general case of polygonal partitions.

\section{Conclusion and Future Work}
\label{sec:conclusion}
We have presented an algebraic approach to the precise and global explanation of Rectifier 
Neural Networks (PLNNs), one of the most popular kinds of Neural Networks. Key
to our approach is the symbolic execution of these networks that allows the construction 
of semantically equivalent \emph{Typed Affine Decision
Structures} (TADS).  Due to their deterministic and sequential nature, TADS can be considered as white-box models and therefore as precise solutions to the 
model explanation problem, which directly imposes also solutions to the outcome explanation, and class characterization problems \cite{gossen2021formal,gossen2021algebraic}. Moreover, as linear algebras, TADS support operations 
that allows one to elegantly 
compare Rectifier Networks for equivalence or $\epsilon$-similarity, both with precise diagnostic 
information in case of failure, and to characterize their classification potential by precisely 
characterizing the set of inputs that are specifically classified, or the set of inputs where two
Network-based classifiers differ. These are steps towards a more rigorous understanding of 
Neural Networks that is required when applying them in safety-critical domains without the 
possibility of human interference, such as self-driving cars.

This elegant situation at the semantic TADS level is in contrast with today's practical reality 
where people directly work with learned PLNNs that are in particular characterized 
by their hidden layers that often comprise millions sometimes even billions of 
parameters. The reason for this complex structure is learning efficiency,  a property
paid for with semantic intractability: There is essentially no way to control the impact 
of minor changes of a parameter or input values, and even the mere evaluation for
a sample input exceed the capacity of a human's mind by far. This is why PLNNs
are considered as black-box models.

The reason why TADS have not yet been studied may be due to their size:
they may be exponentially larger than a corresponding PLNN\@. The reason for this
expansion is the transformation of the incomprehensible hidden layers structure into
a large decision structure, which conceptually is as easy to comprehend as a 
decision tree and a linear classifier. In this sense, our transformation into TADS can be regarded as 
trade of size for transparency, turning the verification and explanation problem into
a scalability issue. There are at least three promising angles for attacking the scalability
problem: 
\begin{enumerate}
	\item Learned PLNNs have a high amount of noise resulting from the underlying learning 
	process that works by locally optimizing individual parameters of the hidden layers. 
	Noise reduction may have a major impact on size. Detecting noise is clearly a semantic 
	task and can therefore profit from TADS-based semantic analyses.
	\item PLNNs are accepted to be approximate. Thus, controlled modifications with minor 
	semantic impact are easily tolerated. TADS provide the means to control the effect of modifications
	and thereby to keep modifications in the tolerable range.
	\item Modern neural network architectures are typically compositions of multiple sub-networks
	that are intended to support the learning of different subtasks. However, this structure
	at the representational layer gets semantically blurred during joint the learning process, which, e.g.,
	prohibits compositional
	approaches as known from formal methods. The semantic transparency 
	of TADS may provide means to reinforce the intended compositional structure also at the
	semantical level in order to support compositional reasoning and incremental construction.
\end{enumerate}
Of course, there seems to be a hen/egg problem here. If we can construct the TADS, we are able to reduce it 
in order to achieve scalability. On the other hand, we need scalability first to construct the TADS\@. 
This is a well-known problem in the formal methods world, and despite a wealth of heuristics and 
domain-specific technologies, the answer is \emph{compositionality} and \emph{incremental construction}.
This is exactly in line with the observation reported in the third item above: We need to learn how to
use divide and conquer techniques for PLNN in a semantics-aware fashion. TADS are designed to 
support this quest by providing both a leading mindset and a tool-supported technology.

\bibliographystyle{alpha}
\bibliography{literature}

\end{document}